\documentclass{article}

\usepackage{preprint,times}
\iclrfinalcopy
\usepackage[utf8]{inputenc}
\usepackage[T1]{fontenc}
\usepackage{url}
\usepackage{graphicx}
\usepackage{flafter}
\usepackage{wrapfig}
\usepackage{needspace}
\usepackage{booktabs}
\usepackage{array}
\usepackage{amsmath}
\usepackage{amssymb}
\usepackage{algorithm}
\usepackage{algpseudocode}
\usepackage[most]{tcolorbox}
\newtcolorbox{promptbox}[1]{breakable,title={#1},
  colback=black!2,colframe=black!35,colbacktitle=black!8,coltitle=black,
  fonttitle=\small\bfseries,fontupper=\small,
  boxrule=0.4pt,arc=1mm,left=2mm,right=2mm,top=2mm,bottom=2mm}
\usepackage{microtype}

\usepackage{xcolor}
\usepackage{enumitem}
\usepackage[colorlinks=true,allcolors=blue!55!black]{hyperref}
\usepackage[capitalize,noabbrev]{cleveref}

\newcommand{\reset}{Permanent Reset}
\newcommand{\algo}{\texorpdfstring{R$^3$}{R3}} %

\title{Absorbed in Inertia: Activation Analysis for Computer-Use Agents}
\author{%
Giulio Segalini\textsuperscript{1}\quad
Zhi Wen Soi\textsuperscript{1}\quad
Jérémie Decouchant\textsuperscript{2}\quad
Lydia Chen\textsuperscript{1,2}\\
\normalfont\textsuperscript{1}Université de Neuchâtel\quad
\normalfont\textsuperscript{2}Delft University of Technology\\
\normalfont\small\texttt{\{giulio.segalini, zhi.soi, yiyu.chen\}@unine.ch}\\
\normalfont\small\texttt{j.decouchant@tudelft.nl}}

\begin{document}
\maketitle

\begin{abstract}
Computer-use agents have become increasingly capable of executing tasks on live desktops through natural-language instructions, based on trajectories of screenshots, actions, and reasoning.
We discover that they can stealthily exhibit \emph{inertia}, in which they repeat fruitless actions despite recognizing that these actions are ineffective.
We hypothesize that inertia is reflected in the agent's internal state, i.e., the activation values of the agent's underlying model, and propose a protocol to measure the relationship between the two.
Extensive analysis of high-dimensional activation states shows that inertia corresponds to an \emph{absorbing region of activation space}, where activation values become stale across actions and even after attempts to steer them.
We conjecture that drastically changing the agents' activations by re-initializing them is necessary to escape inertia.
Specifically, we propose \algo{} (Reset, Reroute, Restore), which temporarily resets the agent's context trajectory to escape the absorbing region and then restores the historical context to effectively complete the task.
Our approach yields 17--55\% lower measured inertia across models relative to unmodified agents.
These results suggest that changing the context can interrupt recurrence more effectively than directly steering the resulting activations.
Our code is available at \url{https://anonymous.4open.science/r/vlm-agent-defense-D076}.
\end{abstract}

\section{Introduction}
\label{sec:introduction}
\begin{figure}[t]
\centering
\includegraphics[width=.9\linewidth]{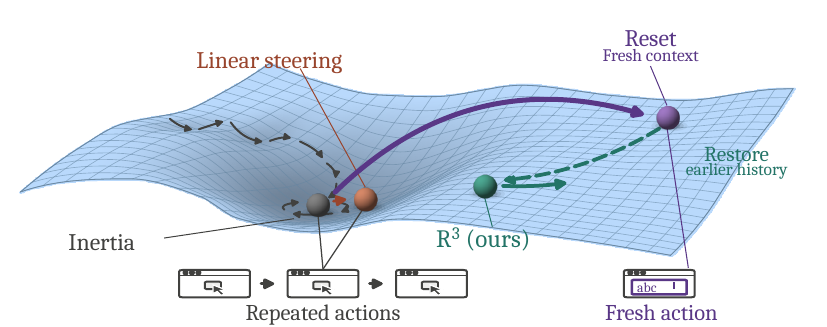}
\caption{\textbf{Activation editing of an absorbing region.}
In inertia, the agent remains near a repeated pre-action state.
Linear steering can preserve this pattern, while context reset recomputes the state from a fresh context and can lead to a different action.
\algo{} restores the earlier history after finding a screen-changing action.
Geometry and arrows are schematic.}
\label{fig:hero}
\end{figure}

A computer-use agent (CUA) turns a natural-language task into mouse and keyboard actions on a live desktop~\citep{qin2025uitars,wang2025opencua,xu2026mobileAgentV35,xue2026evocua}.
Yet a CUA may recognize in its reasoning that its current approach has failed while continuing to issue the same action.
UI-TARS~\citep{qin2025uitars}, for example, writes, ``I just tried using Ctrl+A to select everything, but it didn't work,'' then executes Ctrl+A again from the same screen and leaves it unchanged.
We refer to recurrence that persists despite acknowledged failure as \emph{inertia}.
Unlike broader failure labels based on step repetition~\citep{cemri2025mast}, inertia combines two conditions: an observable lack of progress and an explicit acknowledgment of failure in the agent's reasoning.

Operationally, we detect exact screen--action recurrences that leave the visible state unchanged, then separately assess from the reasoning text whether the agent acknowledges failure.
Across four open models and two computer-use benchmarks, screen--action recurrence is more common in unsuccessful trajectories than in successful ones.
Up to 89\% of screen--action recurrences are accompanied by reasoning that acknowledges the failure.
These cases consume interaction budget even after the agent recognizes that its current approach is ineffective.

We next ask what characterizes the model state that still produces a screen--action recurrence.
We analyze the residual activations immediately before action generation.
Across all tested models, these pre-action activations distinguish screen--action recurrence from other actions.
Among recurrences, they further distinguish cases with and without failure acknowledgment.
Together, these results are consistent with viewing inertia as an absorbing region of activation space, illustrated in Figure~\ref{fig:hero}.

We test whether inertia can be disrupted by linearly steering the pre-action state along directions that distinguish actions with and without screen--action recurrence.
Neither activation steering nor a prompt-level failure notice consistently reduces recurrence.

Rather than perturbing the pre-action state after it has formed, we change the context from which it is computed.
Context reset removes the interaction history, allowing the model to recompute its state from the task and current screen.
At recorded states where the agent has just produced a screen--action recurrence, we compare the next action induced by steering and context reset.
Steering still produces the recurrent action in 69--86\% of samples, whereas context reset does so in at most 4\%.

A context reset alone, however, can discard useful information about task progress and state.
We therefore introduce \algo{} (Reset, Reroute, Restore), which resets the context when recurrent actions leave the screen unchanged, searches for a screen-changing action, and then restores the earlier history with the new action appended.

Our contributions are:

\begin{itemize}[leftmargin=1.1em, topsep=-0.2em, itemsep=2pt, parsep=0pt]

\item \textbf{Measuring and discovering inertia.}
    We identify inertia as recurrence of the same screen--action pair despite failure acknowledgment in the preceding reasoning.
    Across four computer-use agents and two benchmarks, recurrence is more common in unsuccessful trajectories than in successful ones, and up to 89\% of recurrences are accompanied by failure acknowledgment.

\item \textbf{Explaining inertia through activation states.}
    We show that pre-action activations distinguish recurrence from other actions and, among recurrences, inertia from cases without failure acknowledgment.
    These results are consistent with inertia occupying an absorbing region of activation space.

\item \textbf{Mitigating inertia through activation editing.}
    We compare direct linear steering with recomputing the pre-action state through context reset, and introduce \algo{} to restore useful history after finding a screen-changing action.
    \algo{} reduces inertia by 17--55\%, with model-dependent effects on task score.
\end{itemize}

\section{Inertia as an Absorbing Region}
\label{sec:inertia}

We first define how inertia is measured, then examine how it is reflected in the model's internal activations.
\begin{figure}
\centering
\includegraphics[width=\linewidth]{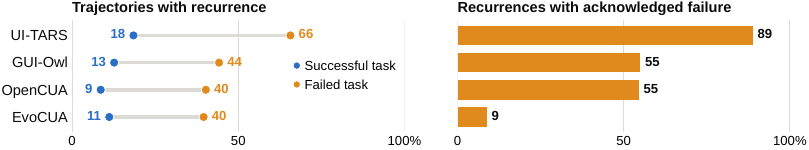}
\caption{\textbf{Recurrence concentrates in failed trajectories.} Recurrence rate (\%) by task outcome (left) and acknowledged share (\%) of recurrences (right) per model. Data from both benchmarks are pooled.}
\label{fig:inertia}
\end{figure}

\subsection{Characterizing Inertia} \label{sec:external_measurement}
\label{sec:definition}

Repeated actions can be useful, for example when an agent scrolls several times to reach the relevant section of a long document.
When scrolling continues to reveal new screens, it does not count as screen--action recurrence.
We use \textbf{screen--action recurrence} as an observable proxy for a lack of \textbf{visible task progress}.
We build a detector that assigns $y_t=1$ when the action at trajectory position $t$ matches one of two patterns: the same action leaves the same screen unchanged across repeated executions, or a sequence of screen--action transitions forms a cycle that returns to its starting screen.
Otherwise, $y_t=0$.
Appendix~\ref{app:tolerance} gives the full detection rules.

To determine whether actions recur despite acknowledged failure, we use a blind LLM judge~\citep{zheng2023llmJudge} to assess the reasoning preceding each detected screen--action recurrence.
We provide only the task and current reasoning, withholding screenshots, executed actions, and detector outputs to keep the assessment independent of recurrence detection.
To keep the judgment separate from recurrence detection, the judge sees only the task and current reasoning, not the screenshots, executed actions, or detector outputs.
We set $f_t=1$ when the reasoning explicitly acknowledges that previous attempts are failing or that the same action recurs without success, and $f_t=0$ otherwise.
Failure acknowledgment alone does not tell us whether an unsuccessful action subsequently recurs.
We define \textbf{inertia} only when both conditions hold, $y_t=f_t=1$.
Appendix~\ref{app:judges} provides the rubric and blinded protocol, and reports validation against different judge models, and human annotations.

\begin{center}
\fbox{
\begin{minipage}{0.88\linewidth}
\textbf{Inertia.}
At trajectory position $t$, inertia occurs when screen--action recurrence and explicit failure acknowledgment coincide:
\[
y_t = f_t = 1.
\]
\end{minipage}
}
\end{center}

Failure acknowledgment can still appear in the reasoning used to justify another attempt.
In a spreadsheet task, EvoCUA~\cite{xue2026evocua} identifies repeated clicks in the middle of the table and states that it needs to change its approach.
It then proposes another click in the same area and executes the same action again, leaving the screen unchanged.
The stated intention to change approach therefore coexists with an unchanged action (Appendix~\ref{app:ack-examples}).

Assessing action utility remains an open problem~\citep{hu2026redundancyBench}.
Existing work assigns broader step-level labels such as redundancy, repetition, unrecoverable failure, or progress~\citep{sui2026tact,cemri2025mast,hu2026redundancyBench,barke2026agentrx,chen2025guiShepherd,zhang2025progrm,xiong2025guiPra,xi2025agentprm,zhang2026dontActBlindly}.
For instance, LLM judges reach 24.88\% step-level F1 against human labels on RedundancyBench~\citep{hu2026redundancyBench}.
Even with few action types, CUAs follow diverse trajectories in which the same action can serve different purposes.
We therefore rely on exact screen--action recurrence, which can be detected mechanically without judging action utility.

We apply our measurement protocol to UI-TARS-1.5-7B~\citep{qin2025uitars}, OpenCUA-32B~\citep{wang2025opencua}, GUI-Owl-1.5-8B-Think~\citep{xu2026mobileAgentV35}, and EvoCUA-32B-S2~\citep{xue2026evocua} on OSWorld~\citep{xie2024osworld} and OpenComputer~\citep{wei2026opencomputer}.
Figure~\ref{fig:inertia} shows that \textbf{recurrence concentrates in unsuccessful trajectories}, occurring in up to 66\% of them.
MAST~\citep{cemri2025mast} likewise identifies step repetition as its most frequent failure mode, accounting for 15.7\% of reported failure-mode occurrences.
Recurrence consumes interaction budget by revisiting previously encountered screens and actions.
For three of the four models, 55--89\% of screen--action recurrences are accompanied by failure acknowledgment in the preceding reasoning.
EvoCUA has the lowest measured rate of inertia among screen--action recurrences.

We next ask whether inertia is reflected in the model's pre-action activations.

\begingroup
\begin{figure}[!t]
\centering
\includegraphics[width=\linewidth]{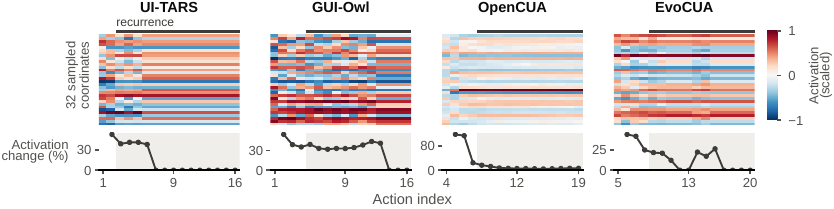}
\caption{\textbf{Activation persistence during recurrence.}
Top: normalized values of 32 sampled pre-action activation coordinates across consecutive actions; the black bar denotes the screen--action recurrence run.
Bottom: change between consecutive activation vectors, measured relative to the model's typical activation magnitude (the median vector norm in its reference sample).
}
\label{fig:native-activation-persistence}
\end{figure}
\endgroup

\subsection{Inertia in Activation Space} \label{sec:internal_measurement} \label{sec:method}

To test whether inertia is reflected in the internal state, we examine the model's activation immediately before action generation, after the reasoning has been produced.
This choice follows naturally from how inertia manifests.
We extract the activation after the reasoning so that its contents are incorporated into the model state, but before the action has been generated.
For the action at trajectory position $t$, we extract the residual-stream activation~\citep{elhage2021transformerCircuits} at layer $\ell$ and denote it $h_t^{(\ell)} \in \mathbb{R}^{d_\ell}$.
This is the same extraction point used by TACT~\citep{sui2026tact}.

\textbf{Activations are stale during recurrence.}
Figure~\ref{fig:native-activation-persistence} shows $h_t^{(\ell)}$ across consecutive actions around the longest screen--action recurrence run for each model.
Before recurrence, the activation coordinates vary across actions.
During recurrence, successive activation vectors change little.
We next test whether pre-action activations distinguish recurrence and failure acknowledgment on held-out tasks.

\begin{wrapfigure}{r}{0.48\textwidth}
\centering
\includegraphics[width=\linewidth]{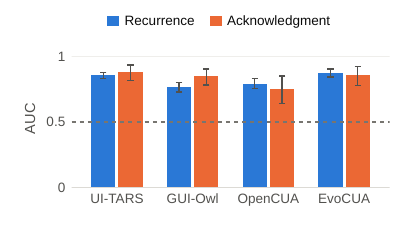}
\caption{\textbf{Activation readouts on held-out tasks.} Mean AUC under nested task validation, with equal task weights. Error bars: 95\% bootstrap intervals. Dashed line: chance.}
\label{fig:activation-readouts}
\end{wrapfigure}

\textbf{Pre-action activations distinguish recurrence and inertia.}
We fit task-balanced activation directions~\citep{sui2026tact,marks2024geometryTruth} and evaluate them on held-out tasks.
For each task containing both classes, we compute the difference between their class-mean activations.
We average these differences with equal weight across tasks, preventing long recurrent sequences from dominating the fitted direction.
The resulting recurrence direction separates actions with and without screen--action recurrence with validation AUC~\citep{fawcett2006roc} of 0.72--0.95 across models.
Among actions with screen--action recurrence, a second direction separates inertia from cases without failure acknowledgment with AUC of 0.82--0.90.
Figure~\ref{fig:activation-readouts} evaluates both directions with nested task validation.
Appendix~\ref{app:activation-geometry} reports the complete AUC results, selected layers, and stability of both directions for each model.

These analyses capture two different properties of inertia.
First, activation persistence measures how much the pre-action state changes over a recurrent sequence.
Second, the readouts ask whether that state carries information about recurrence and failure acknowledgment on held-out tasks.
The results suggest that acknowledged failure remains encoded in the pre-action activations even when those activations change little across recurrent actions.

\section{Activation Editing to Escape Inertia}
\label{sec:activation_editing}
\label{sec:results}
\label{sec:experiments}

\begin{figure}[!ht]
    \centering
    \includegraphics[width=\linewidth]{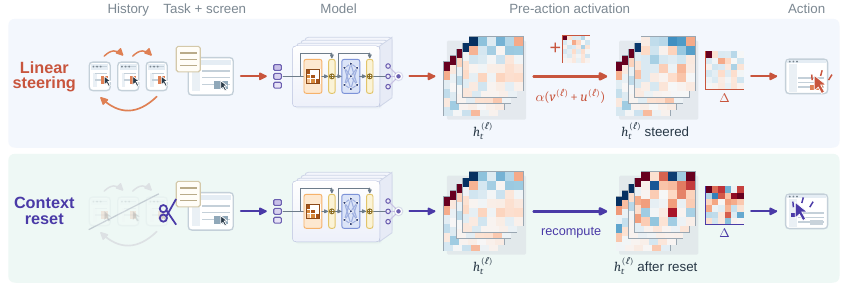}
    \caption{\textbf{Editing pre-action activations.}
    Steering edits the activation, while context reset removes the interaction history before recomputing it. 
    Both rows show measured changes in activations for the two policies.
    Activations matrices are extracted from real data.
    }
    \label{fig:activation-interventions}
\end{figure}

Section~\ref{sec:inertia} links inertia to a persistent pre-action state.
We now test whether changing this state can interrupt inertia.

The model produces action $a_t$ from context $H_t=(x,\tau_{<t},s_t)$, where $x$ is the task instruction, $\tau_{<t}$ is the interaction history, and $s_t$ is the current screenshot.
We write the pre-action activation at layer $\ell$ as $h_t^{(\ell)}=\theta^{(\ell)}(H_t)$.

We compare two ways of changing this state, illustrated in Figure~\ref{fig:activation-interventions}.
Linear steering directly modifies the pre-action activation while keeping the context fixed, $h_t^{(\ell)} \rightarrow h_t^{(\ell)}+\delta^{(\ell)}$.
The second approach changes the context and recomputes the pre-action state from it, $H_t\rightarrow\widetilde H_t\rightarrow\theta^{(\ell)}(\widetilde H_t)$.
We instantiate this approach with \algo{} (Reset, Reroute, Restore).

\subsection{Why Not Just Steering?}
\label{sec:activation-intervention}

We first test whether directly steering the pre-action activations identified in Section~\ref{sec:internal_measurement} can interrupt inertia.

Following TACT~\citep{sui2026tact}, we construct two steering directions from the task-balanced mean differences defined in Section~\ref{sec:method}.
Let $D^{(\ell)}(A,B)$ denote the mean activation difference between classes $A$ and $B$, averaged equally across tasks containing both classes.
The recurrence direction $v^{(\ell)}$ contrasts actions with and without screen--action recurrence.
The inertia direction $u^{(\ell)}$ contrasts inertia with actions without screen--action recurrence, after projecting out the component aligned with $v^{(\ell)}$:
$$
    v^{(\ell)} \propto D^{(\ell)}(y_t=1,\,y_t=0),
    \qquad
    u^{(\ell)} \propto \bigl(I - v^{(\ell)} v^{(\ell)\top}\bigr)\, D^{(\ell)}(y_t=f_t=1,\,y_t=0),
$$
where $\propto$ denotes normalization to unit $\ell_2$ norm.
Unlike the acknowledgment contrast in Section~\ref{sec:method}, $u^{(\ell)}$ uses actions without screen--action recurrence as the negative class.
At each selected layer, we apply
$$
    h_t^{(\ell)}\leftarrow h_t^{(\ell)}
    +\alpha\bigl(v^{(\ell)}+u^{(\ell)}\bigr),
    \qquad \ell\in\mathcal{L},
$$
where $\mathcal{L}$ denotes the selected set of layers and $\alpha$ controls the steering strength.
Under this orientation, $\alpha<0$ subtracts both steering directions.
We use fixed $\alpha$ values of \textminus{}5, \textminus{}1, \textminus{}10, and \textminus{}1 for UI-TARS, GUI-Owl, OpenCUA, and EvoCUA, respectively.
Steering is applied at every action.
Appendices~\ref{app:activation-geometry} and~\ref{app:reset} report the selected layers, steering strengths, and collection protocol.

Under the tested configuration, steering does not consistently reduce inertia or recurrence.
Across the four models, the change in inertia ranges from a 33.8\% reduction to a 44.0\% increase, while the change in recurrence ranges from a 5.6\% reduction to a 39.0\% increase (Table~\ref{tab:policy-compact}).
For each observed reduction, the confidence interval includes zero.
Table~\ref{tab:policy-compact} and Appendix~\ref{app:recurrence-rates} report the full comparison with the other policies.
Figure~\ref{fig:activation-interventions} shows a representative replay in which steering changes the activation but the model still proposes the same click.

This suggests changing the context used to compute the pre-action activation instead of modifying the activation directly.

\subsection{\algo{}: Reset, Reroute, Restore}
\label{sec:mitigations}
Consider an agent that has navigated to a document and then repeatedly presses Ctrl+A without changing the screen.
Reset removes the interaction history, including the failed attempts and the navigation that brought the agent to the document.
Reroute then searches from the task and current screenshot for an action that changes the screen.
Once such an action is found, Restore recovers the history from before the recurrence and appends the accepted action.
The history preceding the recurrence is hidden during Reroute and restored afterward; the recurrent attempts themselves are not.
If Reroute exhausts its search budget without changing the screen, the trajectory terminates.

\textbf{Trigger.}
\algo{} is triggered when the same screen--action pair occurs for $k$ consecutive executions and the screen remains unchanged.
Only single, non-wait actions contribute to this count.
If action $a_t$ triggers the intervention, Reroute begins from the resulting screen $s_{t+1}$.
Appendix~\ref{app:tolerance} specifies screen and action equivalence, while Appendix~\ref{app:reset} gives the counter definition, action exclusions, and intervention limits.

\textbf{Reset.}
When \algo{} is triggered, Reset removes all previous actions, screenshots, and reasoning from the input context.
The model retains only the task instruction $x$ and current screen $s_{t+1}$.
The next pre-action activation is therefore computed without the recurrent sequence that triggered the intervention.
The reset context is
$
    H_{t+1}^{R} = (x,\varnothing,s_{t+1}),
$
from which the model recomputes its pre-action activation as
$
    h_{t+1,R}^{(\ell)} = \theta^{(\ell)}(H_{t+1}^{R}).
$
Section~\ref{sec:extra-exp} further shows that removing only recurrent copies while retaining the remaining recent history is often insufficient to change the next proposed action.

\textbf{Reroute.}
Starting from the reset context $H_{t+1}^{R}$, the model generates new reasoning and proposes actions using only the task instruction and current screenshot.
At attempt $j$, the model executes $a_R^{(j)}$ in the environment and observes the resulting screen $s_R^{(j)}$.
Neither the original trajectory nor earlier Reroute attempts are added to the model context, although all attempts execute sequentially in the same environment.
The first action that produces
$$
s_R^{(j)} \neq s_{t+1}
$$
is accepted, with its action and resulting screen denoted by $a^\star$ and $s^\star$.
We accept an action based only on whether it changes the screen, without judging its task-level progress.
Otherwise, Reroute tries again, for at most $N$ attempts.
If none changes the screen, the trajectory terminates.
The accepted transition $(a^\star,s^\star)$ is then passed to Restore.
Appendix~\ref{app:reset} specifies the temperatures, retry limit, and remaining implementation details.

\textbf{Restore.}
Restore reconstructs the interaction history from before the triggering recurrence and appends the accepted Reroute interaction.
This recovers task information that a permanent reset would discard, including progress made before inertia began.
Let $\tau_R^\star$ denote the accepted Reroute interaction entry, containing its pre-action screenshot, generated reasoning, and action $a^\star$.
Restore reconstructs the history and context as
$$
    \tau^\star = \tau_{<t-k+1} \mathbin{\Vert} \tau_R^\star,
    \qquad H^\star = (x,\tau^\star,s^\star),
$$
where $\tau_{<t-k+1}$ is the history preceding the recurrent sequence that triggered the intervention, $\mathbin{\Vert}$ denotes appending the accepted interaction entry, and $s^\star$ is the current screenshot.
After Restore, subsequent pre-action activations are computed from the pre-recurrence history together with the accepted transition.
The triggering recurrent sequence and rejected Reroute attempts are absent from this context.
Agent execution then continues from $H^\star$.

\section{Evaluation}
\label{sec:protocol}
\label{sec:eval}

\begin{table}[t]
\centering
\caption{\textbf{Intervention effects on OpenComputer.}
Paired changes from the unmodified agent in inertia, screen--action recurrence, and task score, with task-level standard errors after $\pm$.
Inertia and screen--action recurrence are reported as relative changes in their rates per executed action.
Bold and underline mark the best and second-best point estimates per model and metric.}
\label{tab:policy-compact}
\begingroup
\small
\setlength{\tabcolsep}{4pt}
\renewcommand{\arraystretch}{1.1}
\begin{tabular*}{\linewidth}{@{\extracolsep{\fill}}lrrrr@{}}
\toprule
Method & UI-TARS & GUI-Owl & OpenCUA & EvoCUA \\
\midrule
\multicolumn{5}{@{}l}{\textit{Inertia (\%, $\downarrow$)}} \\
\quad Failure notice & $-15.3\pm8.2$ & $+20.5\pm26.8$ & $\underline{-33.1}\pm31.5$ & $-14.4\pm41.9$ \\
\quad Steering & $-7.0\pm8.7$ & $+44.0\pm31.9$ & $+36.7\pm38.7$ & $-33.8\pm31.7$ \\
\quad \reset{} & $\underline{-27.3}\pm7.8$ & $\underline{-6.5}\pm22.8$ & $-9.5\pm37.0$ & $\mathbf{-64.9}\pm30.2$ \\
\quad \textbf{\boldmath\algo{} (ours)} & $\mathbf{-33.9}\pm9.2$ & $\mathbf{-16.5}\pm24.1$ & $\mathbf{-39.2}\pm30.1$ & $\underline{-54.9}\pm33.8$ \\
\midrule
\multicolumn{5}{@{}l}{\textit{Screen--action recurrence (\%, $\downarrow$)}} \\
\quad Failure notice & $-4.6\pm8.4$ & $-5.0\pm16.6$ & $\underline{-23.4}\pm16.9$ & $+16.9\pm17.6$ \\
\quad Steering & $-5.6\pm6.9$ & $-0.2\pm15.7$ & $+39.0\pm19.0$ & $\underline{-0.9}\pm13.0$ \\
\quad \reset{} & $\underline{-14.6}\pm8.0$ & $\underline{-28.8}\pm14.5$ & $-2.3\pm16.7$ & $+6.9\pm18.0$ \\
\quad \textbf{\boldmath\algo{} (ours)} & $\mathbf{-16.9}\pm5.6$ & $\mathbf{-38.0}\pm9.9$ & $\mathbf{-42.7}\pm16.1$ & $\mathbf{-33.8}\pm11.3$ \\
\midrule
\multicolumn{5}{@{}l}{\textit{Task score (points, $\uparrow$)}} \\
\quad Failure notice & $\underline{+1.00}\pm1.51$ & $+0.20\pm2.42$ & $\underline{+2.20}\pm2.91$ & $-3.60\pm2.59$ \\
\quad Steering & $\mathbf{+2.20}\pm2.13$ & $-2.80\pm2.61$ & $\mathbf{+4.20}\pm2.09$ & $\mathbf{-0.60}\pm2.13$ \\
\quad \reset{} & $+0.40\pm1.79$ & $\mathbf{+3.20}\pm2.35$ & $-0.50\pm2.94$ & $-4.60\pm2.75$ \\
\quad \textbf{\boldmath\algo{} (ours)} & $+0.60\pm1.54$ & $\underline{+1.20}\pm1.92$ & $-2.50\pm2.97$ & $\underline{-0.90}\pm1.72$ \\
\bottomrule
\end{tabular*}
\endgroup

\end{table}

We evaluate our approach on the same 97 OpenComputer tasks using four models: UI-TARS-1.5-7B~\citep{qin2025uitars}, OpenCUA-32B~\citep{wang2025opencua}, GUI-Owl-1.5-8B-Think~\citep{xu2026mobileAgentV35}, and EvoCUA-32B-S2~\citep{xue2026evocua}.
We compare \algo{} and \reset{}, which applies only the Reset step before continuing from the fresh context, against the unmodified agent, activation steering, and a prompt-level failure notice that reports how many times the recurrent action left the screen unchanged (Appendix~\ref{app:reset}).
All methods are evaluated over complete trajectories.
Table~\ref{tab:policy-compact} reports paired changes relative to the unmodified agent with task-level standard errors, while the appendix provides 95\% task-bootstrap intervals.

We measure both task performance and the recurrence targeted by the interventions.
Specifically, we report inertia and screen--action recurrence rates, normalized by the number of executed actions to account for differences in trajectory length.
Reset thresholds are chosen from the observed recurrence lengths; Appendix~\ref{app:reset} gives the full selection procedure.

\subsection{Reducing Inertia in Full Trajectories}
\label{sec:intervention-results}

Across complete OpenComputer trajectories, \algo{} \textbf{reduces inertia} for all four models.
At the selected thresholds, inertia decreases by 33.9\%, 16.5\%, 39.2\%, and 54.9\% for UI-TARS, GUI-Owl, OpenCUA, and EvoCUA, respectively.
Screen--action recurrence also decreases by 17--43\%, with all confidence intervals below zero (Appendix~\ref{app:recurrence-rates}).
Table~\ref{tab:policy-compact} compares these changes with the other intervention policies, with full details in Appendix~\ref{app:inertia-rates}.

\algo{} has the largest point-estimate reduction in screen--action recurrence for all four models, while the effects of failure notices, steering, and \reset{} vary across models.
Since \reset{} applies only the Reset step, its comparison with \algo{} shows what is added by Reroute and Restore.
For UI-TARS and GUI-Owl, \reset{} reduces recurrence by 14.6\% and 28.8\%, respectively.
For OpenCUA and EvoCUA, recurrence changes by \textminus{}2.3\% and +6.9\% under \reset{}, compared with reductions of 42.7\% and 33.8\% under \algo{}.
At threshold $k=8$, re-entries per task change by \textminus{}2.85 for OpenCUA and \textminus{}6.16 for EvoCUA under \algo{}, compared with +0.36 and \textminus{}1.27 under \reset{}, respectively (Appendix~\ref{app:reset}).

We report inertia together with screen--action recurrence because inertia can decrease when failure acknowledgment becomes less frequent, even if recurrence does not.
For EvoCUA, \reset{} gives the largest decrease in inertia despite an increase in screen--action recurrence.
Considering inertia alone would therefore suggest an improvement that is not reflected in recurrence.
Under \algo{}, both inertia and screen--action recurrences per action decrease.

\algo{} reduces recurrence more consistently than it improves task score.
Task-score point estimates change by +0.60, +1.20, \textminus{}2.50, and \textminus{}0.90 points for UI-TARS, GUI-Owl, OpenCUA, and EvoCUA, respectively.
Neither negative change has a confidence interval excluding zero.
Recurrence can decrease without improving task completion, since escaping a recurrent action does not guarantee that the rest of the trajectory will complete the task.

With steering, task score and recurrence can move differently.
Steering has the highest task-score point estimate for UI-TARS, OpenCUA, and EvoCUA.
For OpenCUA, task score increases by 4.20 points while screen--action recurrence increases by 39.0\%; both intervals are above zero.

The termination rule also contributes to the reduction in recurrence under \algo{}.
When Reroute exhausts its search budget without changing the screen, the trajectory ends instead of returning to the recurrent action.
For EvoCUA, restoring the trigger history and resuming execution after such a search adds 3.75 re-entries per task into screen--action pairs that had already failed to change the screen (95\% interval [0.36, 7.63]), without a clear task-score gain (Appendix~\ref{app:soundness-controls}).

\subsection{How Reset Breaks Inertia}
\label{sec:extra-exp}

We examine what changes after reset, from the next proposed action to the activations and behavior observed during continued execution.

\textbf{Fresh context reduces repeated proposals.}
We compare interventions by generating action proposals at the same recorded trigger states, without executing them.
After reset, only 0--4\% of proposals match the action that triggered the intervention.
With both steering directions, repetition remains at 69--86\% (Table~\ref{tab:replay-compact} and Appendix~\ref{app:replay}).
We also apply the recurrence direction alone, scaling it to keep the activation edit equally large.
The agent repeats the action in 71--85\% of these draws.
The directions distinguish recurrent actions in Section~\ref{sec:internal_measurement}, but steering along them usually leaves the proposed action unchanged.

Selective reset removes the duplicate attempts while retaining the rest of the history.
We keep the first traversal of the triggering action sequence and delete only its later occurrences.
A note in the prompt names the repeated action.
Even with this edit, the agent proposes that action in 61--74\% of draws (Table~\ref{tab:selective-reset}), compared with the much lower repetition under a full reset.

\begin{table}[!htbp]
\centering
\begin{minipage}{0.75\linewidth}
\caption{\textbf{Same action again (\%).} Composite and recurrence-only steering have matched norms.}
\label{tab:replay-compact}\label{tab:selective-reset}
\begin{tabular*}{\linewidth}{@{\extracolsep{\fill}}lrrrr@{}}
\toprule
& UI-TARS & GUI-Owl & OpenCUA & EvoCUA\\
\midrule
Unmodified & 79.8 & 81.0 & 87.5 & 84.8\\
Composite & 68.5 & 83.3 & 85.7 & 84.8\\
Recurrence-only & 71.0 & 82.1 & 83.9 & 84.8\\
Selective reset & 70.2 & 60.7 & 62.5 & 73.9\\
Reset & 0.8 & 2.4 & 3.6 & 0.0\\
\bottomrule
\end{tabular*}
\end{minipage}
\end{table}

\begin{figure}[b]
\centering
\includegraphics[width=\linewidth]{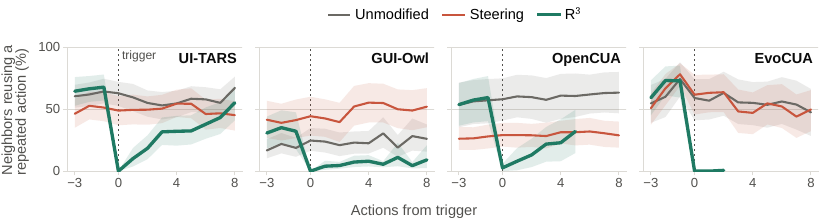}
\caption{\textbf{Activation neighborhoods after the first trigger.}
Share of the 25 nearest unmodified-agent activations associated with reuse of a previously unchanged-screen action.
After reset, this share drops sharply at the trigger.}
\label{fig:neighborhood}
\end{figure}

\textbf{Reset moves activations toward neighborhoods with less action reuse.}
To relate these behavioral changes to the model's internal state, we compare pre-action activations under \algo{} with those of the unmodified agent.
Each activation is assigned its 25 nearest neighbors from other tasks run by the same unmodified model.
We count the fraction of neighbors whose actions reuse a screen--action pair that previously left the screen unchanged.
Lower values mean that fewer neighboring states are associated with these repeated actions.

Figure~\ref{fig:neighborhood} shows a drop in this fraction at reset for every model.
GUI-Owl remains at a low value over the recorded follow-up.
For UI-TARS and OpenCUA, the fraction gradually increases again over subsequent actions.
Fresh contexts may be closest to states from the start of a task, when there are few earlier actions to repeat.
To account for this, we repeat the analysis with reference actions drawn only from the same trajectory position as the query.
Reset still has a lower fraction of neighbors associated with action reuse in every model (Table~\ref{tab:neighborhood-sensitivity}).
Matching action index does not match context length, but the result persists after excluding earlier reference actions.
Unmodified agents retain similar pre-action activations as recurrence continues (Figure~\ref{fig:native-activation-persistence}).
The shift after reset is consistent with an interruption of that pattern.

\begin{figure}[!ht]
\centering
\includegraphics[width=\linewidth]{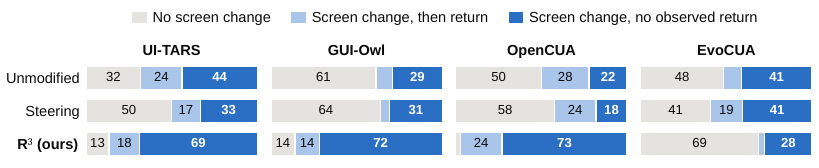}
\caption{\textbf{Screen changes and returns on OpenComputer.}
Percentages of all triggers, averaged over runs.
A return repeats the original screen--action pair after a screen change and leaves the screen unchanged, within recorded follow-up.}
\label{fig:reentry-outcomes}
\end{figure}

\textbf{Screen changes and returns during execution.}
Replay ends with a proposal, before its effect on the screen can be observed.
In full trajectories under \algo{}, we follow execution from each trigger to record whether the screen changes and whether the agent later returns to the original screen--action pair.
For UI-TARS, GUI-Owl, and OpenCUA, \algo{} changes the screen without an observed return to the original screen--action pair in 69\%, 72\%, and 73\% of triggers.
The corresponding values are 44\%, 29\%, and 22\% for the unmodified agent, and 33\%, 31\%, and 18\% under steering (Figure~\ref{fig:reentry-outcomes}).
Some trajectories still return to the original pair within the recorded follow-up.

EvoCUA does not show this improvement during execution.
Reset changes every proposed action in replay, yet screen changes under \algo{} are less frequent than for the unmodified agent (Table~\ref{tab:replay-compact} and Figure~\ref{fig:reentry-outcomes}).
Reroute exhausts its search budget in 19.59\% of trajectories, which then terminate (Appendix~\ref{app:recurrence-rates}).
Stopping these trajectories accounts for part of the reduction in recurrence.
For EvoCUA, restoring the trigger history and continuing after search exhaustion increases exact re-entries without a clear improvement in task score (Section~\ref{sec:intervention-results})

\section{Conclusion}
\label{sec:discussion}

Failure acknowledgment alone does not ensure self-correction in computer-use agents.
We define \emph{inertia} to connect an agent's expressed reasoning to the actions that follow.
We identify recurrence through repeated screen--action pairs.
Through activation analysis, we show that information about failure acknowledgment remains available as the same action recurs.
These observations are consistent with inertia occupying an absorbing region of activation space, where recurrence persists despite detectable failure acknowledgment.
Recovering from inertia therefore requires a more drastic change in the agent's current state, both internally and externally.

Our intervention results show that changing the context can interrupt this recurrence.
Linear steering does not reliably reduce recurrence, whereas \algo{} lets the agent reconsider its next action from a fresh context before restoring the earlier interaction history.
Across four models, \algo{} reduces inertia by 17--55\%, with model-dependent effects on task score.
This temporary reset preserves task context while allowing the pre-action state to be recomputed without the recurrent attempts.
Future work can study which parts of the interaction history should be retained when recovering from failure, which could enable more efficient recovery and lead to higher task-completion rates.

\section*{Reproducibility statement}

The detector compares recorded screens and executed actions under the identity rules in Appendix~\ref{app:tolerance}, including the masked screen regions and the action-normalization keys.
Activation fitting and validation use the disjoint task partitions in Appendix~\ref{app:activation-geometry}, and mitigation comparisons use the paired task-bootstrap intervals in Appendix~\ref{app:reset}.
Appendix~\ref{app:repro} reports the hardware, per-task and per-action elapsed times, and GPU costs of the additional experiments. The released generators record their input sources and hashes.
The failure-acknowledgment judge protocol appears in Appendix~\ref{app:judges}.

\section*{AI use statement}

We used generative AI tools to propose and refine hypotheses, develop and critique research methodology and experiments, and interpret results.
These tools helped implement the exact-recurrence detector, the trajectory normalization and masking pipeline, the judge harness, and the intervention arms.
We also used a text-only judge to annotate failure acknowledgment, as described in Section~\ref{sec:definition}.
We did not use generative AI to generate synthetic datasets, formulate mathematical claims, supply critical ingredients for proofs, assist in writing proofs, or translate material.

We additionally used generative AI to create and modify code to generate scientific figures and images from real data, write and edit software code in the released repository, draft parts of the manuscript, and improve readability.
These tools also supported brainstorming, identifying research topics and gaps, searching for information, finding relevant literature, and summarizing and analyzing prior work.

We take responsibility for the final manuscript and accompanying artifacts, including all content produced with generative AI assistance.

\appendix
\clearpage
\crefalias{section}{appendix}
\crefalias{subsection}{appendix}
\section{Related work}
\label{app:related}

\paragraph{Repetition in text generation.}
Neural text generation repeats itself, a degeneration analyzed theoretically~\citep{fu2021repetitionTheory} and reduced by sampling~\citep{holtzman2020degeneration} or unlikelihood training~\citep{welleck2020unlikelihood}.
\citet{xu2022breakTheLoop} measure the self-reinforcement of repetition, in which each repeated copy raises the probability of the next, and induction heads, which copy from retained context, are a candidate mechanism we do not localize~\citep{olsson2022inductionHeads}.
We test the action-level counterpart, where the repeated segment spans turns with screen observations and persists under the agent's ordinary decoding.

\paragraph{Agent failure taxonomies and step labels.}
Prior work studies reasoning that displaces environment interaction and the recurrence it can produce, including overthinking in agents~\citep{cuadron2025dangerOverthinking}, externally induced reasoning overhead~\citep{kumar2025overthink}, cyclic tool use~\citep{lee2026overthinkingLoops}, failure taxonomies that list recurrence~\citep{cemri2025mast}, semantic step redundancy~\citep{hu2026redundancyBench}, and unrecoverable-failure localization~\citep{barke2026agentrx}.
TACT calls a step overacting when the action returns an observation already in the agent's information state~\citep{sui2026tact}.
A parallel line estimates the usefulness or effect of actions through step-level rewards~\citep{chen2025guiShepherd}, progress prediction~\citep{zhang2025progrm}, visual process evidence~\citep{xiong2025guiPra}, promise-and-progress evaluation~\citep{xi2025agentprm}, and action-effect verification~\citep{zhang2026dontActBlindly}.
Each of these labels is a judgment about the agent's information state or the action's contribution, whereas exact recurrence is a narrower observable assigned directly from screens and actions.
We label it in ordinary trajectories and ask whether it coexists with expressed failure acknowledgment.

\paragraph{Mitigations.}
Reflexion~\citep{shinn2023reflexion} and Self-Refine~\citep{madaan2023selfRefine} organize explicit feedback and revision, whereas we measure acknowledgment and executed actions within the agent's existing reasoning--action sequence.
VLAA-GUI combines completion verification, loop recovery, and on-demand search~\citep{han2026vlaaGui}.
Our failure-notice arm tests a simple prompt-level notice, not this full recovery framework.
Context-management methods manage or consolidate history~\citep{holt2024l2mac,zhou2026mem1} and clear failed attempts before retrying~\citep{arora2026whyRetryingFails}, which motivates \reset{} and \algo{} without training a new memory representation.

\paragraph{Activation steering.}
Activation-steering work fits directions from contrasts and intervenes during generation~\citep{rimsky2024steering,arditi2024refusal,wollschlager2025geometryRefusal}, including for agent exploration~\citep{rahn2024entropicSteering}, multimodal models~\citep{parekh2025learningToSteer}, and coding-agent overthinking and overacting~\citep{sui2026tact}, whose approach we adapt to computer-use agents to test whether activations distinguish actions with screen--action recurrence and whether steering them changes actions.
\par\bigskip
\Needspace{6\baselineskip}
\section{Measurement and annotation}
\label{app:map}
\subsection{Corpus and eligibility}
We analyze 242 OSWorld trajectories per model, except GUI-Owl, which has 241, and 97 OpenComputer trajectories per model (Section~\ref{sec:inertia}).
If a task was run more than once during collection, only the last run with recorded label inputs contributes to the analysis.
OpenComputer trajectories must contain at least one agent action.
Table~\ref{tab:map} reports screen--action recurrence and acknowledgment separately for each model and benchmark.

A trajectory counts as containing screen--action recurrence if at least one action satisfies either detector rule in Appendix~\ref{app:tolerance}.
We report this fraction among all trajectories and separately among those the benchmark marks successful or unsuccessful.
Acknowledgment is measured among actions with screen--action recurrence.
Long repeated sequences contribute more actions to this percentage than short ones. It is not a percentage of tasks.
The successful-trajectory denominators are 58, 79, 78, and 123 on OSWorld and 7, 8, 15, and 29 on OpenComputer, in model order. The two zero rates on successful OpenComputer trajectories therefore represent 0/8 for GUI-Owl and 0/15 for OpenCUA.
\begin{table}[!ht]
\centering\small
\caption{Screen--action recurrence and acknowledgment (\%) by model and benchmark. Recurrence uses eligible trajectories within each outcome group. Acknowledgment uses actions with screen--action recurrence.}
\label{tab:map}
\resizebox{\linewidth}{!}{\begin{tabular}{@{}llrrrr@{}}
\toprule
Model & Benchmark & All trajectories & Failed & Successful & Acknowledged actions \\
\midrule
UI-TARS & OSWorld & 45.9 & 54.3 & 19.0 & 88.1 \\
UI-TARS & OpenComputer & 83.5 & 88.9 & 14.3 & 89.1 \\
GUI-Owl & OSWorld & 29.5 & 37.0 & 13.9 & 36.2 \\
GUI-Owl & OpenComputer & 52.6 & 57.3 & 0.0 & 59.3 \\
OpenCUA & OSWorld & 27.3 & 35.4 & 10.3 & 56.0 \\
OpenCUA & OpenComputer & 42.3 & 50.0 & 0.0 & 51.8 \\
EvoCUA & OSWorld & 20.7 & 33.6 & 8.1 & 11.8 \\
EvoCUA & OpenComputer & 42.3 & 50.0 & 24.1 & 7.6 \\
\bottomrule
\end{tabular}
}
\end{table}

\subsection{Detector implementation and identity sensitivity}
\label{app:tolerance}

\paragraph{Screen and action identity.}
Two screens match ($\equiv$) when all pixels outside the desktop clock region are identical, and two actions match when their command, coordinates, keys, and text are identical.
The screen key $\phi(s)$ is a hash of the image pixels with the desktop clock region masked.
On a $1920\times1080$ screen, the mask covers rows $[0,28)$ and columns $[900,1100)$ on OSWorld, and rows $[0,28)$ and columns $[1690,1920)$ on OpenComputer.
Pixels outside the mask must match exactly. Differences in how the image file is encoded do not affect the comparison.
The mask prevents a clock update from separating otherwise identical screens, but also makes changes inside that region unobservable to the detector.

The action key $\psi(a)$ preserves the executed command and its coordinates, keys, and text.
For cycle matching, we remove formatting differences from GUI commands and compare them in execution order. Reasoning text is excluded.
A block of commands counts as one action and matches only a block with the same commands in the same order.
The detector includes waits and blocks of commands. It excludes actions that only move the cursor and declarations that execution has ended.
Each transition requires both a before-action and an after-action screen. We break the sequence at missing actions or when an action's result screen differs from the screen recorded for the next action.

\paragraph{Recurrence rules.}
Let $a_t$ denote the action at trajectory position $t$ and $s_t$ the screen observed before it.
An eligible transition has signature
\begin{equation}
e_t=\bigl(\phi(s_t),\psi(a_t),\phi(s_{t+1})\bigr).
\end{equation}
For a starting action $i$, a traversal of length $L$ is $C_{i,L}=(e_i,\ldots,e_{i+L-1})$.
At each start, the cycle detector selects the shortest screen-continuous traversal that returns to its initial screen, searching lengths $1\leq L\leq8$.
It groups traversal signatures up to cyclic rotation, then compares non-overlapping occurrences at the same phase.
When several phases recur, it retains the phase with the earliest first and second occurrences.
The first traversal does not count as recurrence. Later matching traversals do.

The unchanged-screen rule also marks an action when the same screen--action key occurred earlier and both executions left that screen unchanged.
An action is labeled repeated if either rule detects it, and is counted only once.
When fitting activation directions, we exclude first traversals from the non-repeated comparison group.

\begin{algorithm}[!ht]
\caption{Exact screen--action recurrence detection}
\label{alg:recurrence}
\small
\begin{algorithmic}[1]
\Require Recorded trajectory, screen key $\phi$, action key $\psi$, cycle bound $L_{\max}=8$
\Ensure Repeated actions $R$, first-traversal actions $F$, eligible actions $E$
\State $R\gets\varnothing$, $F\gets\varnothing$
\State $(E,\mathcal{S})\gets\Call{EligibleSegments}{\text{trajectory},\phi,\psi}$
\ForAll{continuous segments $S\in\mathcal{S}$}
    \State $\mathcal{C}\gets\varnothing$
    \ForAll{starting actions $i$ in $S$}
        \For{$L=1,\ldots,\min(L_{\max},\text{transitions remaining from }i)$}
            \If{$\phi(s_{i+L})=\phi(s_i)$}
                \State Add $C_{i,L}$, its canonical rotation, and its phase to $\mathcal{C}$
                \State \textbf{break} \Comment{Retain the shortest screen return}
            \EndIf
        \EndFor
    \EndFor
    \ForAll{groups $G$ with the same canonical signature in $\mathcal{C}$}
        \State $\mathcal{P}\gets\Call{NonOverlappingOccurrencesByPhase}{G}$
        \State Discard phase lists with fewer than two occurrences from $\mathcal{P}$
        \If{$\mathcal{P}\ne\varnothing$}
            \State $P\gets$ phase list with earliest first and second occurrences
            \State $F\gets F\cup\Call{Actions}{P_1}$
            \State $R\gets R\cup\bigcup_{j=2}^{|P|}\Call{Actions}{P_j}$
        \EndIf
    \EndFor
\EndFor
\ForAll{eligible actions $t\in E$}
    \If{$\phi(s_{t+1})=\phi(s_t)$ and an earlier unchanged execution has the same key}
        \State $R\gets R\cup\{t\}$
    \EndIf
\EndFor
\State \Return $(R,F,E)$
\end{algorithmic}
\end{algorithm}

For example, two consecutive traversals of the same two-screen cycle mark the second traversal.
Two executions of the same hotkey from the same screen, each leaving it unchanged, mark the second execution.
Neither observation establishes that hidden application state stayed fixed or that the action was unnecessary.
The eight-transition search bound limits the cycles this implementation can detect. It is distinct from the reset threshold $k$.
Models can perturb click coordinates while repeating an interaction. Table~\ref{tab:tolerance} therefore tests matching coordinates on a pixel grid, using the same tasks and leaving screen identity unchanged.
Two clicks assigned to the same grid location may target different interface elements, so additional detections are not automatically more accurate.
\begin{table}[!ht]
\centering\small
\caption{Sensitivity of screen--action recurrence to coordinate tolerance. Coordinates are quantized to a $g$-pixel grid on the OpenComputer trajectories in Table~\ref{tab:map}. Detections are counted once per action. Quantized matches can target different interface elements.}
\label{tab:tolerance}
\resizebox{\linewidth}{!}{\begin{tabular}{@{}l rrrr rrrr@{}}
\toprule
 & \multicolumn{4}{c}{trajectories} & \multicolumn{4}{c}{actions} \\
\cmidrule(lr){2-5}\cmidrule(lr){6-9}
Model & exact & $g8$ & $g16$ & $g32$ & exact & $g8$ & $g16$ & $g32$ \\
\midrule
UI-TARS-1.5-7B & 84\% & 84\% & 84\% & 84\% & 3,832 & 3,832 & 3,832 & 3,832 \\
GUI-Owl-1.5-8B-Think & 53\% & 53\% & 54\% & 54\% & 2,158 & 2,191 & 2,216 & 2,223 \\
OpenCUA-32B & 42\% & 46\% & 48\% & 49\% & 732 & 1,007 & 1,099 & 1,177 \\
EvoCUA-32B-S2 & 42\% & 43\% & 43\% & 43\% & 1,441 & 1,445 & 1,449 & 1,460 \\
\bottomrule
\end{tabular}
}
\end{table}

\paragraph{Coordinate sensitivity of mitigation effects.}
We also apply the grid-based action keys to the paired unmodified and \algo{} trajectories, retaining both detector rules and the eight-transition cycle bound.
Table~\ref{tab:tolerance-policy} reports relative changes in screen--action recurrence per executed action. Seeds are averaged within tasks before resampling tasks for the intervals.
The estimated reduction remains negative at every grid size for each model. OpenCUA is more sensitive to coordinate matching, and its intervals include zero at all three relaxed grids.
The other three models retain intervals below zero throughout. Exact matching remains the primary measure, and these checks concern screen--action recurrence rather than acknowledgment on the actions that the relaxed grids add.
\begin{table}[!ht]
\centering\small
\caption{Coordinate sensitivity of \algo{} on OpenComputer. Cells give relative changes in screen--action recurrence per executed action (\%) with 95\% task-bootstrap intervals. Coordinates are quantized to a $g$-pixel grid. GUI-Owl excludes two paired trajectories of one task with no executed-action denominator.}
\label{tab:tolerance-policy}
\resizebox{\linewidth}{!}{%
\begin{tabular}{@{}lrrrr@{}}
\toprule
Model & Exact & $g=8$ & $g=16$ & $g=32$ \\
\midrule
UI-TARS & $-16.9\ [-26.6,-6.1]$ & $-16.3\ [-26.0,-5.6]$ & $-15.9\ [-25.7,-5.0]$ & $-15.4\ [-25.2,-4.7]$ \\
GUI-Owl & $-38.0\ [-52.1,-20.8]$ & $-37.0\ [-51.2,-19.7]$ & $-36.9\ [-50.9,-19.8]$ & $-35.8\ [-50.0,-19.0]$ \\
OpenCUA & $-42.7\ [-61.6,-15.6]$ & $-28.2\ [-49.7,+1.5]$ & $-25.8\ [-47.2,+3.5]$ & $-20.1\ [-43.1,+11.0]$ \\
EvoCUA & $-33.8\ [-47.8,-15.5]$ & $-33.1\ [-47.4,-15.1]$ & $-32.7\ [-46.9,-14.7]$ & $-32.9\ [-47.1,-15.2]$ \\
\bottomrule
\end{tabular}}
\end{table}

\subsection{Failure-acknowledgment protocol and validation}
\label{app:judges}

\paragraph{Rubric and judge inputs.}
The judge assigns a positive label when the reasoning states that the agent's own attempts, actions, or approach are failing or repeating without success.
This includes explicit loop language, an unsuccessful method or target, and an announced retry of an action described as having failed.
Task-required recurrence, neutral continuation, and an earlier difficulty described as resolved without loop language are negative.
Explicit loop language about the agent's own actions remains positive even when the difficulty is described as overcome.
The acknowledgment can refer to a different action from the action executed at that position.
The joint label therefore records co-occurring acknowledgment and recurrence. It does not establish that the emitted action contradicts the plan stated in the reasoning.

The judge receives the task instruction, the current reasoning, and an identifier linking its answer to the action.
It does not receive screenshots, the executed action, or the detector's record of recurrence. Its instructions require judging only the reasoning.
It returns a binary label, a verbatim evidence quote, and a rationale of fewer than 60 words.
The full system prompt appears below; its ``turn'' and ``episode'' denote an action and a trajectory. Its two labels correspond to $f_t=1$ and $f_t=0$ and describe expressed acknowledgment, not latent awareness.
\begin{promptbox}{Failure-acknowledgment judge: system prompt}
You label the pre-action reasoning text of one computer-use agent turn: does the reasoning acknowledge, as its current stance, that the agent's OWN actions are failing or repeating without success? Output exactly one label and a verbatim evidence quote.

- failure\_aware: the text contains a standing claim that its own attempts, actions, or approach are failing, unproductive, or being repeated without success — in any form: explicit repetition or loop language about its own behavior ("I've been going through the same steps repeatedly", "going in circles", "stuck in a loop", "over and over"); a blamed method, target, or outcome ("I've been clicking the color picker, but nothing happens", "the command I entered earlier didn't execute", "I just clicked cell A3, but that's not the right way", "this process several times, and it still doesn't work"); or an announced re-attempt of something that already failed ("click Home again and see if anything is new"). Explicit loop language about its own behavior counts even when the difficulty is described as already overcome.

- unaware: nothing is claimed to be failing about the agent's own actions. This covers: no reference to its own past actions at all; repetition the task itself requires (entering a password twice to confirm it); neutral continuation of an unfinished activity ("the page is long, keep scrolling", waiting more while something loads); a plain earlier difficulty reported as overcome, without loop language ("the first attempts failed, but I finally managed — now ...").

Rules:

- Judge only the reasoning text. Which action the agent actually emitted, how often it was emitted, and what the screen shows are irrelevant; use no outside knowledge of the episode.

- Only the agent's own actions count. A persistent system state ("it is still loading") is not a failure claim unless the agent's own repeated action is also blamed ("I have been waiting and nothing changes" counts; "it is still loading, so I will wait" does not).

- It does not matter WHICH action is acknowledged — the emitted one, a different one, or an unspecified "same steps". Do not try to match the acknowledgment to the emitted action.

- evidence\_quote: copy the exact decisive span verbatim from the reasoning. Required for failure\_aware; for unaware quote the benign mention if one exists, else leave it empty.

- label must be exactly one of: unaware, failure\_aware. Keep the rationale under 60 words. Return one schema-valid JSON object.
\end{promptbox}

\paragraph{Judging and exclusions.}
Judge 1, Qwen3.6-27B-NVFP4, supplies the reported labels. We compare them with labels from Judge 2, Gemma-4-31B-it-NVFP4, to measure agreement.
Both use temperature zero, disabled thinking, and a maximum of 4,096 output tokens. Answers must follow a fixed JSON format. Failed requests are retried, with at most three attempts.
Failed requests are excluded rather than labeled negative. We also exclude a judge's answer if its evidence quote does not appear verbatim in the reasoning.
Agreement is calculated on actions assessed by both judges (Table~\ref{tab:judgeagreement}).
Missing or unreadable reasoning does not count as absence of acknowledgment.

\paragraph{Human validation.}
The human annotator labels 40 actions without seeing the judges' answers. Six actions are repeated without identifying them as retests.
The sample includes four actions both judges labeled negative, their one disagreement, and 35 randomly sampled actions both labeled positive.
Human agreement is 40/40 with Judge 1 and 39/40 with Judge 2, with 6/6 consistent retests.
There are five human negatives from three trajectories and no OpenCUA examples.
Because the sample is selected largely from joint positives, its agreement rates do not estimate population accuracy or positive-label precision for either judge.

A separate blinded check covers 96 unique actions, with 12 from each model--benchmark combination and equal numbers of positive and negative Judge~1 labels within each combination.
Table~\ref{tab:human-balanced} reports agreement with each judge separately by model and benchmark.
The annotator first labels the reasoning without seeing either judge's answer. We retain this blind pass separately from corrections made after comparison with the key.
Three EvoCUA OSWorld acknowledgment labels change from negative to positive on correction. Three other actions change from insufficient to sufficient reasoning without changing their positive acknowledgment labels.
All 96 actions have acknowledgment labels in both passes, and none is marked uncertain.
This check includes negative predictions and OpenCUA, which the first check covers sparsely or not at all. Both checks use one annotator, and their sampling designs do not estimate population accuracy.
\begin{table}[!ht]
\centering\small
\caption{Human agreement on the separately sampled balanced check. Each cell gives agreeing actions out of 12. Blind labels precede comparison with the judge key. Corrected labels are reported separately.}
\label{tab:human-balanced}
\begin{tabular}{@{}llrrrr@{}}
\toprule
& & \multicolumn{2}{c}{Blind pass} & \multicolumn{2}{c}{Corrected pass} \\
\cmidrule(lr){3-4}\cmidrule(l){5-6}
Model & Benchmark & Judge 1 & Judge 2 & Judge 1 & Judge 2 \\
\midrule
UI-TARS & OSWorld & 12/12 & 12/12 & 12/12 & 12/12 \\
UI-TARS & OpenComputer & 11/12 & 11/12 & 11/12 & 11/12 \\
GUI-Owl & OSWorld & 12/12 & 10/12 & 12/12 & 10/12 \\
GUI-Owl & OpenComputer & 11/12 & 12/12 & 11/12 & 12/12 \\
OpenCUA & OSWorld & 10/12 & 10/12 & 10/12 & 10/12 \\
OpenCUA & OpenComputer & 11/12 & 12/12 & 11/12 & 12/12 \\
EvoCUA & OSWorld & 9/12 & 8/12 & 12/12 & 11/12 \\
EvoCUA & OpenComputer & 11/12 & 11/12 & 11/12 & 11/12 \\
\bottomrule
\end{tabular}
\end{table}

\begin{table}[!ht]
\centering\small
\caption{Failure-acknowledgment agreement. Judge 1 supplies the reported labels, which are compared with Judge 2's labels. The note below the table specifies exclusions and the actions used to calculate agreement.}
\label{tab:judgeagreement}
\providecommand{\jafrac}[1]{{\scriptsize\textcolor{black!45}{#1}}}
\begin{tabular}{@{}ll rr rr r@{}}
\toprule
 & & \multicolumn{2}{c}{judged rows} & \multicolumn{2}{c}{acknowledged} & \\
\cmidrule(lr){3-4}\cmidrule(lr){5-6}
Model & Benchmark & all & dropped & judge 1 & judge 2 & agreement \\
\midrule
UI-TARS-1.5-7B & OSWorld & \jafrac{1,597} & \jafrac{0} & 88.1\% & 89.9\% & 97.9\% \\
 & OpenComputer & \jafrac{3,832} & \jafrac{81} & 89.1\% & 90.8\% & 98.1\% \\
\addlinespace[2pt]
GUI-Owl-1.5-8B-Think & OSWorld & \jafrac{505} & \jafrac{1} & 36.2\% & 44.4\% & 91.7\% \\
 & OpenComputer & \jafrac{2,158} & \jafrac{1} & 59.3\% & 59.8\% & 99.5\% \\
\addlinespace[2pt]
OpenCUA-32B & OSWorld & \jafrac{1,323} & \jafrac{0} & 56.0\% & 54.6\% & 97.0\% \\
 & OpenComputer & \jafrac{732} & \jafrac{4} & 51.8\% & 53.3\% & 96.8\% \\
\addlinespace[2pt]
EvoCUA-32B-S2 & OSWorld & \jafrac{561} & \jafrac{0} & 11.8\% & 19.3\% & 92.5\% \\
 & OpenComputer & \jafrac{1,441} & \jafrac{0} & 7.6\% & 12.6\% & 95.0\% \\
\addlinespace[2pt]
\multicolumn{7}{@{}p{\dimexpr\linewidth-2\tabcolsep\relax}@{}}{\scriptsize\color{black!55}\textit{Judged rows are the actions with screen--action recurrence in each cell. A row whose quoted evidence is not verbatim in the reasoning it judges is dropped from that judge's rate and its denominator: judge 1 dropped 1 row and judge 2 dropped 86; the dropped column counts the 87 rows either judge lost, which agreement also excludes.}} \\
\bottomrule
\end{tabular}

\end{table}

\paragraph{Agreement, task weighting, and reasoning exposure.}
\label{app:ack-checks}
Table~\ref{tab:ack-checks} reports Cohen's $\kappa$ on actions assessed by both judges and task-macro acknowledgment for Judge~1.
Macro acknowledgment averages the within-task rate over tasks with at least one judged action with screen--action recurrence. Tasks without such actions are excluded, not assigned zero.
Intervals use 4,000 task-bootstrap draws, retaining all actions from each sampled task. Agreement measures consistency between judges, not accuracy against a human reference.
For positive and negative labels, specific agreement is $2N_{++}/(2N_{++}+N_{+-}+N_{-+})$ and $2N_{--}/(2N_{--}+N_{+-}+N_{-+})$, respectively.
\begin{table}[!ht]
\centering\small
\caption{Task-macro acknowledgment (\%) and judge agreement. Brackets give 95\% task-bootstrap intervals. Specific agreement is reported separately for positive / negative labels (\%).}
\label{tab:ack-checks}
\resizebox{\linewidth}{!}{%
\begin{tabular}{@{}llrrr@{}}
\toprule
Model & Benchmark & Macro acknowledgment & Cohen's $\kappa$ & Specific agreement \\
\midrule
UI-TARS & OSWorld & 76.57 [69.36, 83.23] & 0.892 [0.792, 0.951] & 98.80 / 90.34 \\
UI-TARS & OpenComputer & 87.30 [81.54, 92.60] & 0.898 [0.742, 0.984] & 98.96 / 90.81 \\
GUI-Owl & OSWorld & 45.94 [35.50, 56.24] & 0.828 [0.672, 0.941] & 89.66 / 93.02 \\
GUI-Owl & OpenComputer & 60.32 [49.34, 70.86] & 0.990 [0.980, 0.998] & 99.61 / 99.43 \\
OpenCUA & OSWorld & 55.17 [44.67, 65.16] & 0.939 [0.878, 0.980] & 97.27 / 96.62 \\
OpenCUA & OpenComputer & 50.26 [35.67, 64.59] & 0.937 [0.817, 1.000] & 96.99 / 96.68 \\
EvoCUA & OSWorld & 7.75 [3.26, 13.24] & 0.717 [0.438, 0.851] & 75.86 / 95.57 \\
EvoCUA & OpenComputer & 6.84 [3.24, 10.91] & 0.726 [0.472, 0.912] & 75.17 / 97.22 \\
\bottomrule
\end{tabular}}
\end{table}

At actions with screen--action recurrence, mean exposed reasoning lengths are 74.1, 68.3, 238.3, and 46.3 whitespace-separated words on OSWorld and 75.4, 83.0, 205.2, and 42.9 on OpenComputer, in model order.
No source reasoning field in these corpora is missing or empty under the extraction used here.
Shorter exposed reasoning may express less acknowledgment without implying less internal recognition of failure.

\subsection{Reasoning and executed actions: examples}

\label{app:ack-examples}

These examples illustrate expressed failure acknowledgment alongside recorded screen--action recurrence. They are selected for clarity, rather than to estimate how often a particular form of reasoning occurs. Screen identity uses the clock mask described in Appendix~\ref{app:tolerance}. Quotes reproduce exposed reasoning, with omissions marked by ellipses.

\paragraph{A proposed restart within a repeated cycle (UI-TARS, OpenComputer).}

The task is to generate and export an 880\,Hz sine tone in Audacity. At actions 35--37, the agent produces the same reasoning:

\begin{quote}\small I've found myself stuck in a frustrating loop, repeatedly trying to enter 880Hz without any success. It seems like the issue isn't with my input method, but rather that the input box is just not responding. Instead of continuing to waste time here, it might be better to close this tone settings window and rethink my approach. I need to click the "Cancel" button in the bottom right corner of the window and start over.\end{quote}

\begin{figure}[!ht]

\centering

\includegraphics[width=.85\linewidth]{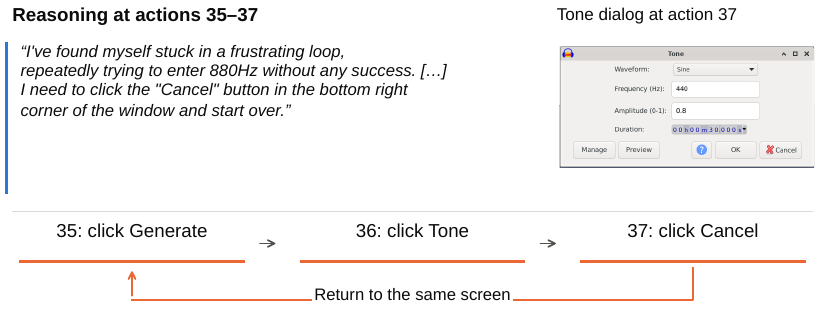}

\caption{\textbf{A repeated three-action cycle under the same failure-acknowledging reasoning.} UI-TARS on OpenComputer, actions 35--37, with the tone dialog as recorded at action 37.}

\label{fig:ack-example}

\end{figure}

The executed actions are clicks on Generate, Tone, and Cancel, respectively, as Figure~\ref{fig:ack-example} shows. These reproduce the screen--action cycle at actions 32--34 and return to the same screen. The screen changes within the cycle. At action 35, the tone dialog is already closed and the click opens Generate, although the reasoning says to click Cancel. At action 37, the click does select Cancel, completing the repeated cycle. This example illustrates both the recurrence and why the acknowledged difficulty must be distinguished from the action executed at that position.

\begin{figure}[!ht]

\centering

\includegraphics[width=\linewidth]{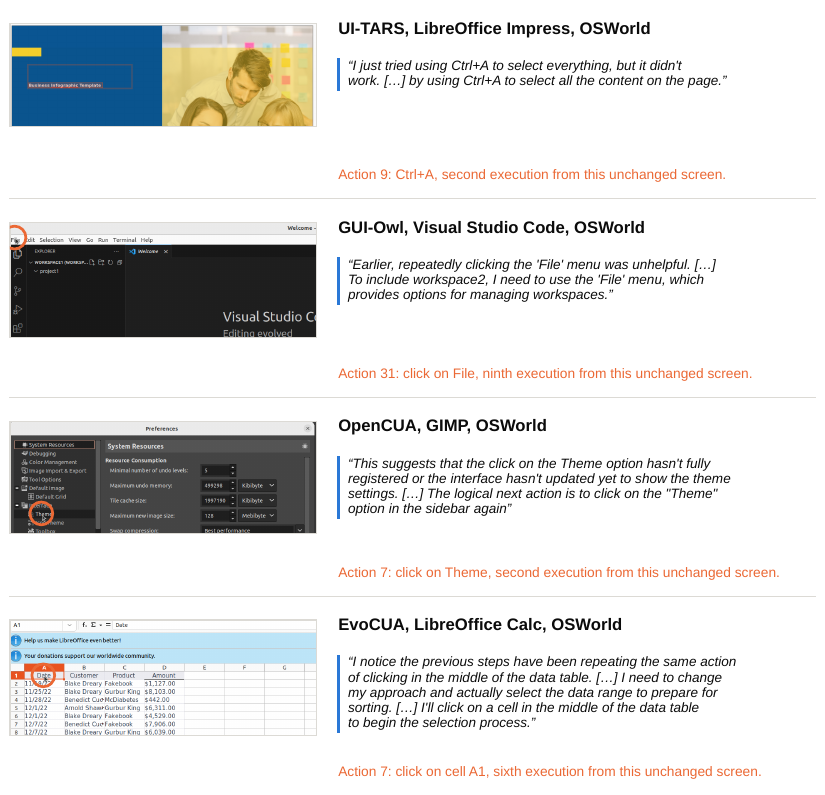}

\caption{\textbf{Failure acknowledgment with a repeated action, one example per model.} Left, the screen at the acknowledged action, with the executed click circled. Right, the exposed reasoning at that action and the executed action, which repeats an earlier execution from the same screen and leaves it unchanged.}

\label{fig:ack-examples}

\end{figure}

Figure~\ref{fig:ack-examples} shows one further example per model, all on OSWorld, in which a single action is repeated from a screen that it leaves unchanged. The four paragraphs below give the reasoning and the executed action of each.

\paragraph{Repeating an unsuccessful shortcut (UI-TARS, OSWorld).}

In a LibreOffice Impress task, the agent writes at action 9:

\begin{quote}\small I just tried using Ctrl+A to select everything, but it didn't work. [\ldots] by using Ctrl+A to select all the content on the page.\end{quote}

It executes Ctrl+A, matching action 8. Both executions start from the same screen and leave it unchanged. The reasoning describes the shortcut as unsuccessful and then proposes using it again.

\paragraph{Reopening a menu after calling the repeated clicks unhelpful (GUI-Owl, OSWorld).}

The task is to open two workspaces in one Visual Studio Code window. At action 31, the agent writes:

\begin{quote}\small Earlier, repeatedly clicking the 'File' menu was unhelpful. [\ldots] To include workspace2, I need to use the 'File' menu, which provides options for managing workspaces.\end{quote}

The executed click is at $(84,69)$, on the File menu, matching 8 earlier actions, most recently 26, 27, and 30. Each execution leaves the same screen unchanged. The reasoning names the repeated clicks as unhelpful and then selects the same menu again.

\paragraph{Retrying an apparently unregistered click (OpenCUA, OSWorld).}

The task is to change GIMP's theme from dark to light. At action 7, the agent writes:

\begin{quote}\small This suggests that the click on the Theme option hasn't fully registered or the interface hasn't updated yet to show the theme settings. [\ldots] The logical next action is to click on the "Theme" option in the sidebar again\end{quote}

The executed click is at $(154,671)$, matching action 6. Both executions leave the same screen unchanged. Here the reasoning acknowledges an apparent failure to register the earlier click and explicitly recommends another click.

\paragraph{Announcing a change of approach and repeating the click (EvoCUA, OSWorld).}

The task is to sort a LibreOffice Calc sheet by amount. At action 7, the agent writes:

\begin{quote}\small I notice the previous steps have been repeating the same action of clicking in the middle of the data table. I need to change my approach and actually select the data range to prepare for sorting. I'll click on a cell in the middle of the data table to begin the selection process.
</think>\end{quote}

The executed click is at $(159,314)$, on cell A1, matching actions 2--6. Each execution leaves the same screen unchanged. The reasoning states that the previous steps repeated the same click, announces a change of approach, and then proposes the same click.

\par\bigskip
\Needspace{6\baselineskip}
\section{Activation extraction, validation, and neighborhoods}
\label{app:activation-geometry}
\subsection{Extraction sites and task split}

We reconstruct the context and reasoning of each labeled action and extract the residual activation after the final token of its reasoning--action marker, before action content.
Table~\ref{tab:extraction} lists the marker used by each model. We match its token sequence to locate the boundary.
Each activation is paired with the label of the same action. Actions without a recorded boundary or activation are excluded.

We split OSWorld tasks into fitting and validation sets while preserving the proportions of application categories. All models and both contrasts use this same split, including the same held-out validation tasks. All actions from one task stay together.
We use the fitting set to estimate directions and the reference projection. We use validation tasks to measure how well the directions distinguish actions with screen--action recurrence and to select intervention layers.
We evaluate interventions on the separate OpenComputer tasks.
Layer indices are one-based. Table~\ref{tab:extraction} lists the captured layer ranges and the ten selected intervention layers.
\begin{table}[!ht]
\centering\small
\caption{Activation extraction sites, captured layer ranges, and steering layers. Layer indices are one-based. Selected layers can be non-contiguous because selection ranks validation AUC rather than layer depth.}
\label{tab:extraction}
\begin{tabular}{@{}lp{.17\linewidth}lp{.44\linewidth}@{}}
\toprule Model & Boundary & Captured layers & Steering layers \\
\midrule
UI-TARS & Action marker & 4--25 & 14, 15, 16, 17, 18, 19, 20, 21, 22, 23 \\
GUI-Owl & Closing \texttt{</think>} & 4--33 & 23, 24, 26, 27, 28, 29, 30, 31, 32, 33 \\
OpenCUA & Action marker & 4--61 & 41, 42, 43, 44, 45, 46, 47, 48, 49, 50 \\
EvoCUA & Opening \texttt{<tool\_call>} & 4--61 & 47, 49, 50, 52, 53, 54, 55, 56, 57, 58 \\
\bottomrule
\end{tabular}
\end{table}

\subsection{Steering-direction construction}
The two contrasts below define the steering directions of Section~\ref{sec:activation-intervention}. The acknowledgment readout in Section~\ref{sec:method} instead contrasts acknowledged and unacknowledged recurrence.
The primary contrast compares actions with screen--action recurrence with eligible actions without screen--action recurrence in the fitting set.
Within each task containing both classes, we subtract the negative-class mean from the positive-class mean.
We average these differences with equal task weight and normalize to obtain $v^{(\ell)}$.

The second contrast uses actions with $y_t=f_t=1$ as positives and eligible actions without screen--action recurrence as negatives.
Actions with screen--action recurrence but without a valid positive acknowledgment do not enter its positive class.
It uses the same task split and averages class differences within tasks, including only tasks with both classes.
Writing its normalized direction as $w^{(\ell)}\propto D^{(\ell)}(y_t=f_t=1,\,y_t=0)$ with the unit-norm $\propto$ of Section~\ref{sec:activation-intervention}, we remove the component shared with the primary contrast:
\begin{equation}
u^{(\ell)}=
\frac{w^{(\ell)}-\langle w^{(\ell)},v^{(\ell)}\rangle v^{(\ell)}}
{\|w^{(\ell)}-\langle w^{(\ell)},v^{(\ell)}\rangle v^{(\ell)}\|_2}.
\end{equation}
Table~\ref{tab:contrast-support} counts the fitting tasks containing both classes for each contrast. These counts can differ because some tasks contain actions with screen--action recurrence but none that is acknowledged. The fitting and held-out task sets remain the same.
\begin{table}[!ht]
\centering\small
\caption{Contributing fitting tasks and actions for the two contrasts on OSWorld. Both contrasts use the same registered task split. A fitting task contributes when it contains both the corresponding positive class and eligible actions without screen--action recurrence.}
\label{tab:contrast-support}
\resizebox{\linewidth}{!}{\begin{tabular}{@{}llrrr@{}}
\toprule
Model & Positive class & Contributing fitting tasks & Positive actions & Negative actions \\
\midrule
UI-TARS & Screen--action recurrence & 75 & 1134 & 1730 \\
UI-TARS & Inertia & 68 & 1020 & 1553 \\
GUI-Owl & Screen--action recurrence & 55 & 397 & 1957 \\
GUI-Owl & Inertia & 36 & 168 & 1262 \\
OpenCUA & Screen--action recurrence & 57 & 1019 & 1127 \\
OpenCUA & Inertia & 43 & 624 & 877 \\
EvoCUA & Screen--action recurrence & 39 & 483 & 943 \\
EvoCUA & Inertia & 12 & 62 & 218 \\
\bottomrule
\end{tabular}
}
\end{table}

\subsection{Validation}
The action recurrence and acknowledgment rates in Figure~\ref{fig:inertia} pool by summing their numerators and denominators across OSWorld and OpenComputer, separately for each model. The reported recurrence AUC uses OSWorld alone.
At each captured layer, validation AUC compares projection scores within tasks containing both classes, then averages across tasks.
We steer the ten layers with the highest validation AUC. Validation tasks are held out from direction fitting and also select these layers, so the reported AUCs are measured on the tasks that select them.

We assess primary-direction stability by fitting directions to 200 task splits and recording their cosine alignment.
For comparison, we shuffle labels within tasks 1,000 times. Each shuffle is summarized by the median split-half cosine using the same task splits.
Table~\ref{tab:readout-support} reports validation AUC and the observed and shuffled-label stability summaries.
We compare the observed alignment with the shuffled-label results without imposing a fixed cosine cutoff.

\begin{table}[!ht]
\centering\small
\caption{Validation AUC and primary-direction stability. Stability entries give the median [minimum, maximum]. The null summarizes median split-half cosines under within-task label permutations. Validation also selects the layer.}
\label{tab:readout-support}
\label{tab:decodable}
\resizebox{\linewidth}{!}{\begin{tabular}{@{}lrrcc@{}}
\toprule
Model & Layer & Val. AUC & Split-half cosine & Shuffled-label null \\
\midrule
UI-TARS & 20 & 0.826 & 0.963 [0.942, 0.974] & -0.002 [-0.215, 0.398] \\
GUI-Owl & 33 & 0.721 & 0.307 [0.121, 0.436] & 0.003 [-0.173, 0.389] \\
OpenCUA & 48 & 0.952 & 0.822 [0.579, 0.888] & -0.005 [-0.436, 0.685] \\
EvoCUA & 56 & 0.930 & 0.817 [0.708, 0.885] & 0.000 [-0.289, 0.363] \\
\bottomrule
\end{tabular}
}
\end{table}

\begin{figure}[!ht]
\centering
\includegraphics[width=.92\linewidth]{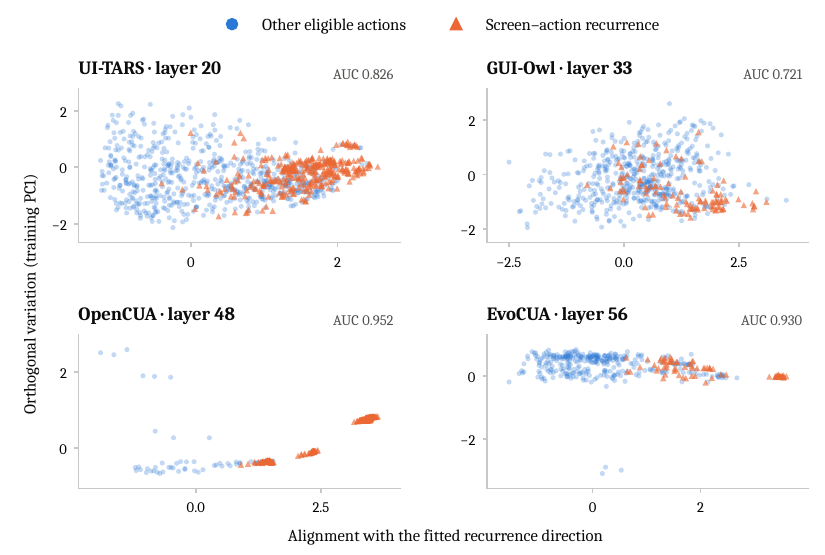}
\caption{Validation actions projected onto the fitted direction $v^{(\ell)}$ and orthogonal training PC1, in training standard deviations. AUC averages within-task comparisons on the data used to select layers. Density reflects action counts.}
\label{fig:activation-context}
\label{fig:activation-pca}
\end{figure}

\paragraph{Projected class overlap.}
Figure~\ref{fig:activation-context} projects validation activations onto the fitted direction $v^{(\ell)}$ and an orthogonal principal component estimated from fitting data.
Both axes are expressed in fitting-data standard deviations.
The density plot weights actions, whereas validation AUC averages within-task comparisons.

\subsection{Inertia readout}
\label{app:inertia-auc}
\begin{figure}[!ht]
\centering
\includegraphics[width=.6\linewidth]{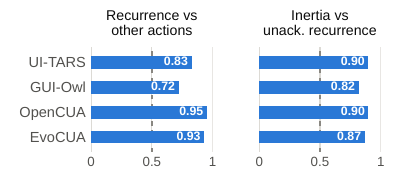}
\caption{Validation AUC of the recurrence readout (left, OSWorld) and the inertia readout (right, both benchmarks). The dashed line marks chance.}
\label{fig:auc-axes}
\end{figure}

The inertia readout of Section~\ref{sec:internal_measurement} contrasts acknowledged ($y_t=f_t=1$) with unacknowledged ($y_t=1$, $f_t=0$) actions with screen--action recurrence, using the within-task fitting procedure of the recurrence direction.
Figure~\ref{fig:auc-axes} places it next to the recurrence readout.
Each benchmark is split into fitting and validation tasks with the same seed and selects its own layer by validation AUC.
The reported value averages task-level AUCs with equal weight over the validation tasks of both benchmarks that contain both classes: 0.899 for UI-TARS (22 tasks), 0.818 for GUI-Owl (31), 0.899 for OpenCUA (14), and 0.869 for EvoCUA (22).
The same tasks select the layer, so these are validation rather than test values.
The paragraphs below give the results for each benchmark and for the registered OSWorld split.

\paragraph{OpenComputer.}
The acknowledgment contrast among actions with screen--action recurrence is repeated on the unmodified agents' OpenComputer trajectories, with the judge labels of the same instrument and the boundary activations of every executed action at every layer.
OpenComputer uses one seeded half split of the 97-task roster, shared by all models, for fitting and validation.
UI-TARS has 3,771 acknowledged and 128 unacknowledged actions with screen--action recurrence, 16 fitting and 11 validation tasks with both classes, a task-macro AUC of 0.976 (sd 0.033) at layer 22, and a split-half cosine of 0.383 against a label-shuffled maximum of 0.231.
OpenCUA has 391 and 400 such actions, 3 fitting and 4 validation tasks with both classes, an AUC of 0.776 (sd 0.248, per-task range 0.43 to 1.00) at layer 49, and a split-half cosine of 0.183 that 6\% of label-shuffled fits exceed.
GUI-Owl has 1,039 acknowledged and 1,431 unacknowledged such actions, 21 fitting and 20 validation tasks with both classes, an AUC of 0.809 (sd 0.320, per-task range 0.00 to 1.00, with 13 of the 20 tasks at or above 0.95) at layer 11, 0.795 at the layer its OSWorld contrast selected, and a split-half cosine of 0.707 against a label-shuffled maximum of 0.323.
EvoCUA has 292 acknowledged and 1,244 unacknowledged such actions, 8 fitting and 13 validation tasks with both classes, an AUC of 0.954 (sd 0.060, per-task range 0.83 to 1.00) at layer 47, and a split-half cosine of 0.053 inside the label-shuffled range (maximum 0.459), so its direction is not shown to be stable.

\paragraph{Seeded OSWorld split.}
Because the registered split leaves OpenCUA and EvoCUA without an evaluable validation task, the OSWorld contrast is repeated on a seeded half split of the 242 tasks, sorted and shuffled once with the OpenComputer seed, 121 fitting and 121 validation tasks shared by all models, with every other setting unchanged.
UI-TARS has 15 fitting and 11 validation tasks with both classes, a task-macro AUC of 0.822 (sd 0.288) at layer 21, and a split-half cosine of 0.327 that 0.2\% of label-shuffled fits exceed.
GUI-Owl has 7 and 11 such tasks, an AUC of 0.834 (sd 0.188) at layer 21, and a split-half cosine of 0.098 that 5\% of label-shuffled fits exceed.
OpenCUA has 10 and 10 such tasks, an AUC of 0.949 (sd 0.158, per-task range 0.50 to 1.00) at layer 49, and a split-half cosine of 0.073 that 38\% of label-shuffled fits exceed, so its direction is not shown to be stable.
EvoCUA has 3 and 9 such tasks, an AUC of 0.747 (sd 0.353, per-task range 0.12 to 1.00) at layer 42, and a split-half cosine of 0.189 that 0.3\% of label-shuffled fits exceed.

\paragraph{Registered OSWorld split.}
The same contrast can also be fitted on the registered split used for the recurrence direction, with the same labels and fitting procedure.
The frozen validation split contains tasks with both classes for UI-TARS (1,369 acknowledged and 150 unacknowledged such actions, seven validation tasks) and GUI-Owl (183 and 322, six tasks), and zero or one such task for OpenCUA and EvoCUA, so those two models are not evaluable on this split and the reported inertia AUC uses the seeded splits instead.
At the validation-selected layer, task-macro AUC is 0.836 for UI-TARS (task-level range 0.50--1.00) and 0.951 for GUI-Owl (0.75--1.00), with split-half cosines of 0.388 and 0.101 against label-shuffled nulls whose 75th percentiles are 0.052 and 0.039.
The acknowledgment direction differs from $v^{(\ell)}$, with a cosine of 0.67 for UI-TARS and 0.02 for GUI-Owl.

\subsection{Activation neighborhoods}
\label{app:neighborhood}
Steering neighborhoods use captures from the selected-strength full trajectories.
We match each response to its capture through the request receipt.
For UI-TARS, GUI-Owl and EvoCUA, we disambiguate identical response texts by requiring all recorded generation events for a request to fall between the recorded interaction's start and its first action's start, after converting collector timestamps to the worker clock.
We retain only unique matches and require that no request is assigned to more than one recorded interaction.
Responses without an action-start timestamp and OpenCUA responses require a unique response-text hash match.
Responses with ambiguous matches or without a boundary capture are excluded, so the steering curves describe the available captures rather than every action.
Hash-only matching can select against repeated responses and does not yield a random sample of steering actions.
Figure~\ref{fig:neighborhood} leaves gaps at offsets with fewer than ten contributing trajectories, following the same display-support rule for all policies.
EvoCUA's curve ends too early to show persistence after the trigger.
The context reset in \algo{} removes the interaction history, so a shift toward states resembling the start of a task is consistent with the intervention's design.
At the first trigger, 96.5\% of UI-TARS's \algo{} neighbors and 100\% of GUI-Owl's and EvoCUA's are from the first two actions of reference tasks. The corresponding share for OpenCUA is 28.0\%.
Table~\ref{tab:neighborhood-sensitivity} restricts the reference library to actions at the same trajectory index as the query, still excluding the query task and taking 25 nearest neighbors.
The repeated-action-neighbor share rises after this restriction but remains below the unmodified share in each model.
Matching action index does not match prompt length. No first-trigger query has 25 reference activations from other tasks with exactly the same prompt-token count.
The neighborhood change therefore describes the state induced by fresh context, without separating context length from recurrence-specific representation changes. The two policies also contribute different query states rather than paired counterfactual activations.
\begin{table}[!ht]
\centering\small
\caption{Activation-neighborhood sensitivity at the first trigger. Cells give the percentage of 25 nearest reference neighbors associated with reuse of an unchanged-screen action. The restricted library matches trajectory action index, not prompt length.}
\label{tab:neighborhood-sensitivity}
\begin{tabular}{@{}lrrrr@{}}
\toprule
& \multicolumn{2}{c}{Original library} & \multicolumn{2}{c}{Same-index library} \\
\cmidrule(lr){2-3}\cmidrule(l){4-5}
Model & Unmodified & \algo{} & Unmodified & \algo{} \\
\midrule
UI-TARS & 62.8 & 0.0 & 48.7 & 30.5 \\
GUI-Owl & 24.7 & 0.0 & 31.3 & 18.5 \\
OpenCUA & 58.0 & 2.7 & 26.5 & 17.7 \\
EvoCUA & 59.0 & 0.0 & 29.8 & 16.5 \\
\bottomrule
\end{tabular}
\end{table}
\par\bigskip
\Needspace{6\baselineskip}
\section{Mitigation protocols and executed outcomes}
\label{app:reset}
\subsection{Policy specifications}
\label{app:steering}
\paragraph{Linear steering.}
At every reasoning--action boundary, we add $\alpha(v^{(\ell)}+u^{(\ell)})$ at each selected layer in Table~\ref{tab:extraction}.
This is the position at which the directions are fitted and at which TACT~\citep{sui2026tact} applies its steering.
This update is applied at every action, regardless of the detector and judge labels.
Its continuous schedule differs from the event-triggered reset policies.
The strength multiplies the sum of the two unit directions directly. We do not rescale this sum or divide by the activation standard deviation.
The tested strength is held fixed throughout a trajectory.

\paragraph{Failure notice.}
After two consecutive executions of the same action leave the screen unchanged, the notice policy adds a factual message to the task instruction for the next model call. It keeps the interaction history.
Its template is:
\begin{quote}\small
You are resuming this task; [steps] steps have been taken so far. Verified interaction record: [action] was performed on the current screen [count] times; the screen did not change.
\end{quote}
The action field contains the recorded action, and the count is rendered as ``twice'' when it equals two.
The message does not replace the task, block an action, or use a judge verdict.

\paragraph{\reset{} and \algo{}.}
Both controllers count consecutive single, non-wait actions with the same screen--action key whose executions leave the screen unchanged.
Unlike the measurement detector, this trigger does not detect multi-action cycles whose intermediate screens change.
Let $c_t$ denote this count after executing action $a_t$, with actions indexed from $t=1$ and $c_0=0$.
Let $s_t$ and $s_{t+1}$ denote the screenshots before and after execution, and let $\equiv$ denote the action and screenshot matching relations from Appendix~\ref{app:tolerance}.
For actions included in the count, the update is
$$
    c_t =
    \begin{cases}
        0, & \text{if $s_{t+1}\not\equiv s_t$},\\
        c_{t-1}+1, & \text{otherwise, if $a_t\equiv a_{t-1}$ and $s_t\equiv s_{t-1}$},\\
        1, & \text{otherwise}.
    \end{cases}
$$
For $t=1$, comparisons with the preceding action and screen are false.
Thus, the first counted action that leaves the screen unchanged starts a new run with $c_t=1$.
After executing $a_t$, the controller triggers when $c_t$ reaches $k$, with $s_{t+1}$ as the current screen, subject to the intervention limits below.

At threshold $k$, \reset{} discards the interaction history and continues from the task and current screen.
\algo{} saves the history and samples up to three proposals from fresh contexts, executing them sequentially in the same environment.
We compare each attempt's result with the screen at the trigger. The environment is not restored between attempts.
After a screen-changing attempt, it restores the history preceding the entire triggering sequence and appends the accepted attempt and its result.
The triggering sequence, including its first occurrence, and rejected search attempts are excluded from the restored history.
Screen change is the acceptance criterion, not a check of task progress. An unchanged-screen attempt can still alter hidden application state, and removing its record from the restored history does not undo that effect.
Execution ends if all search attempts fail to change the screen. \reset{} instead continues from its fresh history even if its first action leaves the screen unchanged.
The comparison therefore tests the combined effects of context reset, live search, history restoration, and termination.

\paragraph{Selective reset.}
Selective reset removes repeated copies of the triggering sequence, retains its first traversal and other history, and adds a note identifying the repeated action.
Appendix~\ref{app:replay} compares its next-action proposals with those from fresh context at the same recorded states.

\label{app:controller-settings}
Both reset policies allow one intervention per screen--action key and at most three per trajectory. \reset{} samples one proposal. \algo{} allows three proposals, using temperature 0.2 for the first and 1.0 for retries.
These intervention limits were chosen to bound additional computation.

\subsection{Parameter settings}
We select steering strengths on the OSWorld validation tasks, which are held out from direction fitting and separate from the OpenComputer evaluation tasks.
Table~\ref{tab:strength-selection} reports every tested strength.
One rule applies to all four models: among strengths whose paired task-score change is within 5 points of the unmodified agent, we select the one with the largest paired reduction in screen--action recurrence.
The rule selects $\alpha=-5,-1,-10,-1$ for UI-TARS, GUI-Owl, OpenCUA, and EvoCUA, respectively, and the full-trajectory and replay comparisons use these strengths.
The largest tested magnitude, $\alpha=-20$, lowers task score by 16.0 and 7.3 points in UI-TARS and GUI-Owl, which the rule excludes.

\begin{table}[!ht]
\centering\small
\caption{Steering-strength selection on the OSWorld validation tasks. Task score and its paired change are in points out of 100. The recurrence change is the paired change in screen--action recurrences per trajectory on the tasks common to all strengths. Evaluable counts tasks with a recorded score.}
\label{tab:strength-selection}
\resizebox{\linewidth}{!}{\begin{tabular}{@{}llrrrrrc@{}}
\toprule
Model & $\alpha$ & Task score & $\Delta$ score & $\Delta$ screen--action recurrence & Mean actions & Evaluable & Selected \\
\midrule
UI-TARS & Unmodified & 28.00 & 0.00 & 0.00 & 35.88 & 50 &  \\
UI-TARS & $0.5$ & 22.00 & -6.00 & +0.14 & 33.40 & 50 &  \\
UI-TARS & $-0.5$ & 19.25 & -8.75 & +0.50 & 34.66 & 50 &  \\
UI-TARS & $-1$ & 26.00 & -2.00 & +0.98 & 33.40 & 50 &  \\
UI-TARS & $-2$ & 26.00 & -2.00 & -0.26 & 37.38 & 50 &  \\
UI-TARS & $-5$ & 24.00 & -4.00 & -3.08 & 31.48 & 50 & $\checkmark$ \\
UI-TARS & $-10$ & 26.00 & -2.00 & -2.26 & 35.14 & 50 &  \\
UI-TARS & $-20$ & 12.00 & -16.00 & +15.24 & 49.54 & 50 &  \\
GUI-Owl & Unmodified & 31.55 & 0.00 & 0.00 & 26.54 & 49 &  \\
GUI-Owl & $-0.5$ & 28.25 & -10.03 & -1.47 & 26.58 & 49 &  \\
GUI-Owl & $-1$ & 32.61 & -1.71 & -1.39 & 28.50 & 49 & $\checkmark$ \\
GUI-Owl & $-2$ & 37.37 & +4.11 & -0.33 & 28.20 & 48 &  \\
GUI-Owl & $-5$ & 34.24 & +1.33 & -0.19 & 27.14 & 47 &  \\
GUI-Owl & $-10$ & 31.71 & +1.08 & -0.58 & 30.80 & 41 &  \\
GUI-Owl & $-20$ & 27.08 & -7.26 & -1.89 & 28.18 & 48 &  \\
OpenCUA & Unmodified & 32.69 & 0.00 & 0.00 & 26.66 & 50 &  \\
OpenCUA & $-1$ & 32.00 & -0.70 & +1.66 & 27.54 & 50 &  \\
OpenCUA & $-2$ & 34.00 & +1.30 & -0.50 & 26.54 & 50 &  \\
OpenCUA & $-5$ & 30.00 & -2.70 & +0.54 & 28.26 & 50 &  \\
OpenCUA & $-10$ & 38.00 & +5.30 & -0.70 & 24.76 & 50 & $\checkmark$ \\
EvoCUA & Unmodified & 46.82 & 0.00 & 0.00 & 28.92 & 50 &  \\
EvoCUA & $-1$ & 45.97 & -0.85 & -0.24 & 28.36 & 50 & $\checkmark$ \\
EvoCUA & $-2$ & 48.94 & +2.12 & +0.22 & 29.48 & 50 &  \\
EvoCUA & $-5$ & 45.43 & -1.40 & +0.66 & 29.88 & 50 &  \\
EvoCUA & $-10$ & 51.97 & +5.15 & -0.10 & 28.42 & 50 &  \\
\bottomrule
\end{tabular}
}
\end{table}

The reported \algo{} thresholds are $k=8$ for UI-TARS and $k=4$ for GUI-Owl, OpenCUA, and EvoCUA. The permanent \reset{} baseline uses $k=8$ for all four models. Table~\ref{tab:reset} reports the tested \algo{} settings $k\in\{4,8\}$.
These choices use the reported OpenComputer comparisons, not a separate held-out threshold-selection set. Selection favors the largest reduction in re-entries among settings whose paired task-score interval includes zero, or the largest task-score change if no setting meets that condition.
An interval containing zero does not establish equivalent task performance. Using the same tasks for selection and reporting can favor the selected estimates.

\subsection{Full results}
For each model, we compare intervention and unmodified trajectories of the same task and seed under matching collection settings.
We calculate the difference in re-entry counts, defined below, and in task score for each pair, then average across seeds within each task.
The reported mean and 95\% bootstrap interval use tasks as the statistical unit.
Task score is expressed in points out of 100. Models and benchmarks are not pooled.
Table~\ref{tab:policy-comparison} includes raw re-entry totals and the number of paired trajectories as well as task-level changes.
Each task has equal weight in the reported mean, even when the number of paired trajectories differs between tasks.
\begin{table}[!ht]
\centering\small
\caption{Paired policy outcomes on OpenComputer. Re-entry columns count screen--action re-entries, as defined in Appendix~\ref{app:recurrence-rates}, for the unmodified agent / intervention. This counter differs from the full detector used in Table~\ref{tab:policy-compact}. Changes are paired means with 95\% task-bootstrap intervals. Each policy uses all valid matched trajectories at its specified settings, so seed coverage differs across rows. Search thresholds are the reported per-model thresholds. Steering uses the fixed strength in Appendix~\ref{app:controller-settings}.}
\label{tab:policy-comparison}
\label{tab:family-steering}
\resizebox{\linewidth}{!}{\begin{tabular}{@{}llrrrr@{}}
\toprule
Model & Policy & Pairs & Re-entry totals & $\Delta$ re-entries / task [95\% CI] & $\Delta$ score [95\% CI] \\
\midrule
UI-TARS & Failure notice & 97 & 2113 / 2432 & $+3.29\ [-3.30, +9.71]$ & $+1.00\ [-2.10, +3.80]$ \\
UI-TARS & \reset{} ($k=8$) & 97 & 2113 / 1426 & $-7.08\ [-12.46, -1.81]$ & $+0.40\ [-3.20, +3.60]$ \\
UI-TARS & \algo{} ($k=8$) & 193 & 4795 / 2607 & $-11.24\ [-15.78, -6.96]$ & $+0.60\ [-2.50, +3.40]$ \\
UI-TARS & Linear steering ($\alpha=-5$) & 97 & 2682 / 2335 & $-3.58\ [-10.51, +3.69]$ & $+2.20\ [-1.90, +6.50]$ \\
GUI-Owl & Failure notice & 97 & 1516 / 1544 & $+0.29\ [-6.88, +7.62]$ & $+0.20\ [-4.50, +4.90]$ \\
GUI-Owl & \reset{} ($k=8$) & 97 & 1516 / 834 & $-7.03\ [-13.61, -0.54]$ & $+3.20\ [-1.20, +7.80]$ \\
GUI-Owl & \algo{} ($k=4$) & 194 & 3226 / 1292 & $-9.97\ [-14.52, -5.60]$ & $+1.20\ [-2.60, +4.80]$ \\
GUI-Owl & Linear steering ($\alpha=-1$) & 97 & 1710 / 1908 & $+2.04\ [-5.91, +10.02]$ & $-2.80\ [-8.10, +2.10]$ \\
OpenCUA & Failure notice & 97 & 885 / 697 & $-1.94\ [-5.40, +1.59]$ & $+2.20\ [-3.60, +7.90]$ \\
OpenCUA & \reset{} ($k=8$) & 97 & 885 / 920 & $+0.36\ [-3.15, +4.07]$ & $-0.50\ [-6.30, +5.10]$ \\
OpenCUA & \algo{} ($k=4$) & 97 & 885 / 524 & $-3.72\ [-7.09, -0.54]$ & $-2.50\ [-8.30, +3.20]$ \\
OpenCUA & Linear steering ($\alpha=-10$) & 95 & 694 / 1081 & $+4.07\ [+0.41, +7.97]$ & $+4.20\ [+0.20, +8.40]$ \\
EvoCUA & Failure notice & 95 & 943 / 1144 & $+2.12\ [-2.26, +6.43]$ & $-3.60\ [-8.90, +1.40]$ \\
EvoCUA & \reset{} ($k=8$) & 97 & 943 / 820 & $-1.27\ [-5.91, +3.35]$ & $-4.60\ [-10.30, +0.50]$ \\
EvoCUA & \algo{} ($k=4$) & 194 & 2050 / 642 & $-7.26\ [-11.45, -3.64]$ & $-0.90\ [-4.10, +2.50]$ \\
EvoCUA & Linear steering ($\alpha=-1$) & 97 & 1107 / 1184 & $+0.79\ [-2.97, +4.49]$ & $-0.60\ [-5.00, +3.40]$ \\
\bottomrule
\end{tabular}
}
\end{table}

\paragraph{Screen--action recurrence per executed action.}
\label{app:recurrence-rates}
The re-entry counter in Tables~\ref{tab:policy-comparison} and \ref{tab:reset} counts a single non-wait action when its screen--action key previously left the screen unchanged, even if its current execution changes the screen.
It is distinct from the full repeat-and-cycle detector in Appendix~\ref{app:tolerance}.
We reconstruct that detector from adjacent recorded screenshots preceding actions and divide its count by executed actions within each trajectory, then average paired changes across seeds within tasks and across tasks (Table~\ref{tab:recurrence-rates}).
Command blocks count as one action. Terminal declarations and controller-stop records are excluded from the denominator. A final action without a following screenshot remains in the denominator but cannot be labeled.
The thresholds are not reselected on these rates.
Table~\ref{tab:policy-compact} reports relative changes in inertia and in this rate as the mean paired change and its task-level standard error, both divided by the unmodified task mean.
\begin{table}[!ht]
\centering\small
\caption{Screen--action recurrence per executed action on OpenComputer under \algo{}. Rates are task-macro percentages. Changes give percentage points with 95\% task-bootstrap intervals. Stops are explicit controller terminations after search exhaustion.}
\label{tab:recurrence-rates}
\begin{tabular}{@{}lrrrr@{}}
\toprule
Model & Unmodified & \algo{} & Change [95\% CI] & Stops / trajectories \\
\midrule
UI-TARS & 42.78 & 35.56 & $-7.22\ [-11.85,-2.66]$ & 10 / 193 \\
GUI-Owl & 24.54 & 15.21 & $-9.33\ [-14.23,-4.55]$ & 9 / 194 \\
OpenCUA & 20.41 & 11.70 & $-8.71\ [-15.29,-2.63]$ & 1 / 97 \\
EvoCUA & 17.19 & 11.39 & $-5.81\ [-9.75,-2.14]$ & 38 / 194 \\
\bottomrule
\end{tabular}
\end{table}

Relative reductions, computed as one minus the ratio of \algo{} to unmodified task-macro rates, are 16.87\%, 38.02\%, 42.69\%, and 33.78\%, respectively.
Using detector-eligible transitions instead gives rate changes of \textminus{}7.21, \textminus{}9.54, \textminus{}8.73, and \textminus{}5.82 percentage points, respectively, with all four intervals below zero.
Rate normalization does not remove the effect of termination on which actions are observed. Controller termination occurs in 5.18\%, 4.64\%, 1.03\%, and 19.59\% of \algo{} trajectories, respectively.

\paragraph{Inertia per executed action.}
\label{app:inertia-rates}
The inertia block of Table~\ref{tab:policy-compact} counts actions with screen--action recurrence whose reasoning the judge of Appendix~\ref{app:judges} labels as acknowledging failure, divided by executed actions, and pairs each intervention trajectory with the judged unmodified trajectory of the same task.
The judge receives the eligible single actions of Appendix~\ref{app:tolerance}, so repeated waits and command blocks count as screen--action recurrence but not as inertia.
Table~\ref{tab:policy-intervals} gives 95\% task-bootstrap intervals for every policy in Table~\ref{tab:policy-compact}.
The \algo{} point estimates are inertia reductions in all four models, but the intervals are wide. Only the UI-TARS interval excludes zero, so the size of the inertia reduction is uncertain for GUI-Owl, OpenCUA, and EvoCUA.
The corresponding reductions in actions with screen--action recurrence have intervals below zero in all four models.
A lower joint-label rate can also reflect less expressed acknowledgment without a corresponding reduction in repeated actions.
\begin{table}[!ht]
\centering\small
\caption{Relative changes from the unmodified agent reported in Table~\ref{tab:policy-compact}, with 95\% task-bootstrap intervals. Both measures are rates per executed action on OpenComputer.}
\label{tab:policy-intervals}
\begin{tabular}{@{}llrrr@{}}
\toprule
Model & Policy & Pairs & Inertia (\%) [95\% CI] & Screen--action recurrence (\%) [95\% CI] \\
\midrule
UI-TARS & Failure notice & 97 & $-15.3\ [-29.0, +0.9]$ & $-4.6\ [-19.3, +13.9]$ \\
UI-TARS & Steering & 97 & $-7.0\ [-22.4, +11.6]$ & $-5.6\ [-18.0, +8.8]$ \\
UI-TARS & \reset{} & 97 & $-27.3\ [-39.2, -13.4]$ & $-14.6\ [-27.9, +0.9]$ \\
UI-TARS & \algo{} & 193 & $-33.9\ [-47.2, -17.2]$ & $-16.9\ [-26.3, -6.6]$ \\
GUI-Owl & Failure notice & 97 & $+20.5\ [-23.1, +102.1]$ & $-5.0\ [-32.3, +33.2]$ \\
GUI-Owl & Steering & 97 & $+44.0\ [-12.3, +150.9]$ & $-0.2\ [-27.0, +34.9]$ \\
GUI-Owl & \reset{} & 97 & $-6.5\ [-40.4, +53.5]$ & $-28.8\ [-48.2, -2.1]$ \\
GUI-Owl & \algo{} & 194 & $-16.5\ [-51.3, +39.1]$ & $-38.0\ [-52.1, -20.9]$ \\
OpenCUA & Failure notice & 97 & $-33.1\ [-72.3, +44.0]$ & $-23.4\ [-47.6, +11.4]$ \\
OpenCUA & Steering & 95 & $+36.7\ [-29.4, +186.6]$ & $+39.0\ [+0.8, +97.0]$ \\
OpenCUA & \reset{} & 97 & $-9.5\ [-60.5, +104.8]$ & $-2.3\ [-28.7, +37.4]$ \\
OpenCUA & \algo{} & 97 & $-39.2\ [-69.2, +29.8]$ & $-42.7\ [-61.1, -15.9]$ \\
EvoCUA & Failure notice & 95 & $-14.4\ [-73.0, +100.1]$ & $+16.9\ [-14.5, +65.1]$ \\
EvoCUA & Steering & 97 & $-33.8\ [-73.2, +41.9]$ & $-0.9\ [-22.7, +28.6]$ \\
EvoCUA & \reset{} & 97 & $-64.9\ [-82.9, -22.3]$ & $+6.9\ [-22.9, +53.0]$ \\
EvoCUA & \algo{} & 194 & $-54.9\ [-84.8, +14.9]$ & $-33.8\ [-48.1, -15.1]$ \\
\bottomrule
\end{tabular}

\end{table}

\subsection{Threshold sensitivity and controls}
Table~\ref{tab:reset} retains the tested thresholds, including alternatives to the selected policy.
These comparisons show how re-entries and task score change together as intervention timing changes.

\begin{table}[!ht]
\centering\small
\caption{\algo{} results by threshold. Pairs count trajectories of the same task and seed under matching collection settings.}
\label{tab:reset}
\resizebox{\linewidth}{!}{\begin{tabular}{@{}lrrrrc@{}}
\toprule
Model & $k$ & Pairs & $\Delta$ re-entries [95\% CI] & $\Delta$ score [95\% CI] & Selected \\
\midrule
UI-TARS & 8 & 193 & $-11.24\ [-15.78, -6.96]$ & $+0.60\ [-2.50, +3.40]$ & $\checkmark$ \\
UI-TARS & 4 & 97 & $-7.53\ [-13.64, -1.44]$ & $+0.20\ [-3.80, +4.10]$ &  \\
GUI-Owl & 8 & 97 & $-8.79\ [-15.08, -2.76]$ & $+1.80\ [-2.30, +6.30]$ &  \\
GUI-Owl & 4 & 194 & $-9.97\ [-14.52, -5.60]$ & $+1.20\ [-2.60, +4.80]$ & $\checkmark$ \\
OpenCUA & 8 & 97 & $-2.85\ [-6.37, +0.68]$ & $+2.80\ [-2.70, +8.30]$ &  \\
OpenCUA & 4 & 97 & $-3.72\ [-7.09, -0.54]$ & $-2.50\ [-8.30, +3.20]$ & $\checkmark$ \\
EvoCUA & 8 & 97 & $-6.16\ [-10.85, -2.10]$ & $-6.60\ [-12.40, -1.20]$ &  \\
EvoCUA & 4 & 194 & $-7.26\ [-11.45, -3.64]$ & $-0.90\ [-4.10, +2.50]$ & $\checkmark$ \\
\bottomrule
\end{tabular}
}
\end{table}

\paragraph{Soundness controls.}
\label{app:soundness-controls}
We vary the proposal history and the exhaustion rule separately, holding the task roster, threshold, three-attempt cap, sampling temperatures, screen-change acceptance rule, successful-history restoration, and interaction budget fixed.
Changes below count exact re-entries per task and use paired 95\% task-bootstrap intervals.
At the reported thresholds, retaining the full trigger history rather than using fresh history changes exact re-entry by \textminus{}0.604~[\textminus{}4.896, 3.552] for UI-TARS and +1.351~[\textminus{}3.144, 5.825] for GUI-Owl; the corresponding task-score intervals include zero.
After search exhaustion, restoring the trigger history and continuing rather than stopping changes exact re-entry by \textminus{}0.521~[\textminus{}5.250, 4.031], +0.515~[\textminus{}2.258, 3.577], and +3.753~[0.361, 7.629] for UI-TARS, GUI-Owl, and EvoCUA, respectively; all three task-score intervals include zero.
Only EvoCUA's re-entry interval excludes zero, indicating that stopping after exhausted search contributes to its lower re-entry count.
The continuation control removes failed proposal turns and restores the trigger history without rolling back the environment.

\subsection{Executed persistence and return}
\label{app:continuations}
\paragraph{Event selection and definitions.}

For \algo{}, observation starts when a recorded reset fires. We check the first reset by recomputing the trigger from the recorded actions and screens.
For steering, observation starts where the same reset rule would fire. Steering is active throughout the trajectory, including before this point.
We use the reported thresholds, $k=8$ for UI-TARS and $k=4$ for GUI-Owl, OpenCUA, and EvoCUA, with the controller's limits on resets per key and per trajectory.

The reference for each event is the screen, with its clock masked, and the single action at that point.
Persistence counts further executions of that same action from that same screen that leave it unchanged, before the first screen change.
A screen change requires an executed action followed by a screen different from the original screen.
A return requires the original action to execute again from the original screen after that change, with its screen before the next action unchanged.
Revisiting the screen or proposing the original action is insufficient.

Observation continues until the recorded trajectory ends and includes later \algo{} interventions.
Declarations of success or failure do not count as executed actions.
We cannot verify a return if the final action has no following screenshot. Missing screenshots or gaps in the recorded actions cause the analysis to stop with an error.
Screen-change rates use all events. Return rates use the events where the screen changed, including those where execution ended before another action.

\paragraph{Continuation results.}
Table~\ref{tab:continuation} reports screen change and verified return after triggers, separately by model, seed, and policy. Steering uses the fixed strength. These rates describe what happens after a trigger, whereas Table~\ref{tab:policy-compact} measures actions with screen--action recurrence over the full trajectory.
Steering cells at a strength other than the selected one are omitted, and seeds are not pooled.
Intervals use 4,000 task-bootstrap draws within each model, policy, and seed, keeping all events from a sampled task together.
Resamples with no screen changes do not define a conditional return rate and are omitted from that interval.
The reported follow-up distribution counts executed actions after the screen change.
The policies can reach different states and have different numbers of actions left. These rates therefore do not compare interventions from identical starting states.
\begin{table}[!ht]
\centering\small
\caption{Executed screen changes and subsequent recurrence on OpenComputer. Cells give counts, percentages, and 95\% task-bootstrap intervals. Return rates use only events with a screen change. Follow-up gives median [minimum, maximum] executed actions after screen change. Each policy is measured on its own trajectories.}
\label{tab:continuation}
\resizebox{\linewidth}{!}{\begin{tabular}{@{}lclccc@{}}
\toprule
Model & Seed & Policy & \shortstack{Screen\\changes} & \shortstack{Original recurrence\\resumes} & \shortstack{Follow-up\\actions} \\
\midrule
UI-TARS & 0 & \algo{} & \shortstack{32/37 \\ 86.5 [73.5, 97.1]} & \shortstack{5/32 \\ 15.6 [4.2, 28.1]} & 46 [2, 88] \\
UI-TARS & 1 & \algo{} & \shortstack{39/45 \\ 86.7 [76.2, 95.6]} & \shortstack{10/39 \\ 25.6 [11.9, 43.2]} & 33 [0, 84] \\
UI-TARS & default & Linear steering ($\alpha=-5$) & \shortstack{18/36 \\ 50.0 [31.2, 68.4]} & \shortstack{6/18 \\ 33.3 [12.5, 57.1]} & 44.5 [0, 76] \\
GUI-Owl & 0 & \algo{} & \shortstack{28/33 \\ 84.8 [71.9, 96.0]} & \shortstack{4/28 \\ 14.3 [3.4, 28.0]} & 32.5 [2, 91] \\
GUI-Owl & 1 & \algo{} & \shortstack{27/31 \\ 87.1 [72.4, 97.1]} & \shortstack{5/27 \\ 18.5 [5.6, 33.3]} & 48 [7, 92] \\
GUI-Owl & default & Linear steering ($\alpha=-1$) & \shortstack{14/39 \\ 35.9 [18.9, 52.5]} & \shortstack{2/14 \\ 14.3 [0.0, 33.3]} & 67.5 [2, 94] \\
OpenCUA & 0 & \algo{} & \shortstack{18/19 \\ 94.7 [81.2, 100.0]} & \shortstack{5/18 \\ 27.8 [9.1, 53.8]} & 24 [7, 42] \\
OpenCUA & 1 & \algo{} & \shortstack{18/18 \\ 100.0 [100.0, 100.0]} & \shortstack{4/18 \\ 22.2 [0.0, 42.1]} & 27 [0, 53] \\
OpenCUA & default & Linear steering ($\alpha=-10$) & \shortstack{14/33 \\ 42.4 [24.2, 61.8]} & \shortstack{8/14 \\ 57.1 [30.0, 81.8]} & 9.5 [0, 73] \\
EvoCUA & 0 & \algo{} & \shortstack{10/26 \\ 38.5 [19.2, 57.7]} & \shortstack{1/10 \\ 10.0 [0.0, 27.8]} & 54.5 [2, 80] \\
EvoCUA & 1 & \algo{} & \shortstack{7/29 \\ 24.1 [0.0, 44.1]} & \shortstack{1/7 \\ 14.3 [0.0, 50.0]} & 58 [0, 70] \\
EvoCUA & default & Linear steering ($\alpha=-1$) & \shortstack{19/32 \\ 59.4 [41.4, 75.0]} & \shortstack{6/19 \\ 31.6 [9.5, 57.1]} & 46 [0, 90] \\
\bottomrule
\end{tabular}
}
\end{table}

\paragraph{First-trigger sensitivity.}
Table~\ref{tab:first-trigger} uses only the first trigger in each trajectory, so each trajectory contributes at most one event.
\begin{table}[!ht]
\centering\small
\caption{Screen change and return using only the first trigger in each trajectory. Rates and intervals follow Table~\ref{tab:continuation}; follow-up counts executed actions after screen change.}
\label{tab:first-trigger}
\resizebox{\linewidth}{!}{\begin{tabular}{@{}lrlrrr@{}}
\toprule
Model & Seed & Policy & Screen change (\%) & Return (\%) & Median follow-up \\
\midrule
UI-TARS & seed 0 & R$^3$ & 24/28 (85.7 [71.4, 96.4]) & 4/24 (16.7 [4.0, 33.3]) & 53.5 \\
UI-TARS & seed 1 & R$^3$ & 29/33 (87.9 [75.8, 97.0]) & 9/29 (31.0 [15.2, 50.0]) & 41 \\
UI-TARS & default & Linear steering ($\alpha=-5$) & 15/28 (53.6 [35.7, 71.4]) & 5/15 (33.3 [9.1, 58.8]) & 45 \\
GUI-Owl & seed 0 & R$^3$ & 19/21 (90.5 [76.2, 100.0]) & 4/19 (21.1 [5.0, 41.2]) & 40 \\
GUI-Owl & seed 1 & R$^3$ & 18/22 (81.8 [63.6, 95.5]) & 3/18 (16.7 [0.0, 35.7]) & 51 \\
GUI-Owl & default & Linear steering ($\alpha=-1$) & 11/27 (40.7 [22.2, 59.3]) & 1/11 (9.1 [0.0, 30.0]) & 81 \\
OpenCUA & seed 0 & R$^3$ & 13/14 (92.9 [78.6, 100.0]) & 4/13 (30.8 [7.7, 57.1]) & 26 \\
OpenCUA & seed 1 & R$^3$ & 12/12 (100.0 [100.0, 100.0]) & 1/12 (8.3 [0.0, 25.0]) & 27 \\
OpenCUA & default & Linear steering ($\alpha=-10$) & 12/26 (46.2 [26.9, 65.4]) & 6/12 (50.0 [21.4, 78.6]) & 9.5 \\
EvoCUA & seed 0 & R$^3$ & 8/23 (34.8 [17.4, 56.5]) & 0/8 (0.0 [0.0, 0.0]) & 56 \\
EvoCUA & seed 1 & R$^3$ & 3/25 (12.0 [0.0, 24.0]) & 1/3 (33.3 [0.0, 100.0]) & 67 \\
EvoCUA & default & Linear steering ($\alpha=-1$) & 14/24 (58.3 [37.5, 79.2]) & 5/14 (35.7 [11.8, 61.5]) & 54 \\
\bottomrule
\end{tabular}
}
\end{table}

\subsection{Computational resources and intervention costs}
\label{app:repro}

We ran inference with vLLM on one NVIDIA RTX PRO 6000 Blackwell GPU (96~GB), with an Intel Core Ultra 7 265K CPU and 64~GB of host memory.
Table~\ref{tab:gpu-cost} reports mean elapsed time per task and total elapsed time divided by executed actions for the unmodified agents on OpenComputer.
These times include model inference and environment interaction and verification. Concurrent task durations are not GPU reservation hours.

\begin{table}[!ht]
\centering\small
\caption{Elapsed time for unmodified OpenComputer agents. Command blocks count as one action; terminal declarations are excluded from the action count.}
\label{tab:gpu-cost}
\begin{tabular}{@{}lrr@{}}
\toprule
Model & Seconds / action & Hours / task \\
\midrule
UI-TARS & 7.1 & 0.17 \\
GUI-Owl & 14.3 & 0.25 \\
OpenCUA & 240.4 & 2.26 \\
EvoCUA & 71.8 & 1.27 \\
\bottomrule
\end{tabular}

\end{table}

The additional full-history search controls used 3.82 GPU-hours, exhaustion-continuation controls 11.21 GPU-hours, corrected-steering failure-acknowledgment judging 0.75 GPU-hours, and matched-state replay 0.45 GPU-hours.
These are single-GPU wall-clock costs for the recorded collection or reservation windows.

\paragraph{Trajectory length and intervention counts.}

Table~\ref{tab:policy-costs} reports total trajectory length and intervention frequency for the two reset policies. Length counts trajectory records, including terminal records, whereas Table~\ref{tab:policy-comparison} counts re-entries.
\algo{} executes 121, 93, 23, and 140 search actions, including 39, 30, 4, and 85 retries, respectively, over the paired trajectories in Table~\ref{tab:recurrence-rates}.
Mean paired elapsed-time changes are +67, \textminus{}93, \textminus{}992, and \textminus{}572 seconds in model order. These include environment and serving time and do not isolate compute cost.
\begin{table}[!ht]
\centering\small
\caption{Reset-policy trajectory length and intervention frequency on OpenComputer. Length is the paired change in trajectory records per task, including terminal records, with a 95\% interval. Intervention totals and affected trajectory counts include the seeds used in the paired comparison.}
\label{tab:policy-costs}
\resizebox{\linewidth}{!}{\begin{tabular}{@{}llrrr@{}}
\toprule
Model & Policy & $\Delta$ total records [95\% CI] & Interventions & Trajectories affected \\
\midrule
UI-TARS & \reset{} ($k=8$) & $-2.09\ [-7.68, +3.24]$ & 44 & 31 \\
UI-TARS & \algo{} ($k=8$) & $-5.29\ [-9.37, -1.41]$ & 82 & 61 \\
GUI-Owl & \reset{} ($k=8$) & $-4.08\ [-10.55, +2.29]$ & 23 & 16 \\
GUI-Owl & \algo{} ($k=4$) & $-8.84\ [-13.51, -4.17]$ & 64 & 43 \\
OpenCUA & \reset{} ($k=8$) & $+5.00\ [+0.59, +9.54]$ & 19 & 11 \\
OpenCUA & \algo{} ($k=4$) & $+0.90\ [-2.73, +4.60]$ & 19 & 14 \\
EvoCUA & \reset{} ($k=8$) & $+3.08\ [-2.05, +8.36]$ & 25 & 18 \\
EvoCUA & \algo{} ($k=4$) & $-9.08\ [-13.75, -4.43]$ & 55 & 48 \\
\bottomrule
\end{tabular}
}
\end{table}
\par\bigskip
\Needspace{6\baselineskip}
\section{Paired next-action replay}
\label{app:replay}
\subsection{Replay protocol}
For replay, we reconstruct recorded OpenComputer actions from the task, current screen, and interaction history.
Replay covers all four models.
The displays use each task's first recorded reset trigger at the reported threshold: $k=8$ for UI-TARS and $k=4$ for GUI-Owl, OpenCUA, and EvoCUA. Tasks without such a trigger do not enter this replay comparison.
The unmodified and steered conditions use the same generated reasoning up to the action boundary.
We place the boundary at the last prompt token and apply steering there before generating the action.
Reset generates new reasoning from the task and current screen.
Reset therefore changes the history and reasoning, while steering keeps them fixed.

For each recorded state, we sample two reasoning sequences and two actions per sequence.
These samples are grouped by their original state and task. They are not independent trajectories.
We use the model's action parser to check whether each generated response contains an executable action.
Replay proposals are not executed. Context reset omits \algo{}'s retries and history restoration.

For each model, we fit a fixed PCA basis on the OSWorld activations used to fit the direction $v^{(\ell)}$, giving each task equal total weight. We project all OpenComputer intervention conditions into this basis without refitting it. Captures occur after each layer's steering update and include effects propagated from earlier layers. Table~\ref{tab:projection-variance} reports the explained variance of the reference projection.

A \emph{proposal escape} is a parser-valid proposal that differs from the repeated action. Proposal escapes, repeated proposals, and parse failures use all proposals as the denominator. A proposal escape does not establish an executed screen change or task progress.

\begin{table}[!ht]
\centering\small
\caption{Explained variance of the two-dimensional reference projection used in Figure~\ref{fig:activation-movement}.}
\label{tab:projection-variance}
\begin{tabular}{@{}lrrr@{}}
\toprule
Model & Layer & PC1 variance (\%) & PC2 variance (\%) \\
\midrule
UI-TARS & 20 & 14.08 & 4.82 \\
GUI-Owl & 33 & 11.59 & 5.13 \\
OpenCUA & 48 & 13.17 & 8.75 \\
EvoCUA & 56 & 10.46 & 7.36 \\
\bottomrule
\end{tabular}

\end{table}

\subsection{Proposal outcomes}
\begin{figure}[!ht]
\centering
\includegraphics[width=\linewidth]{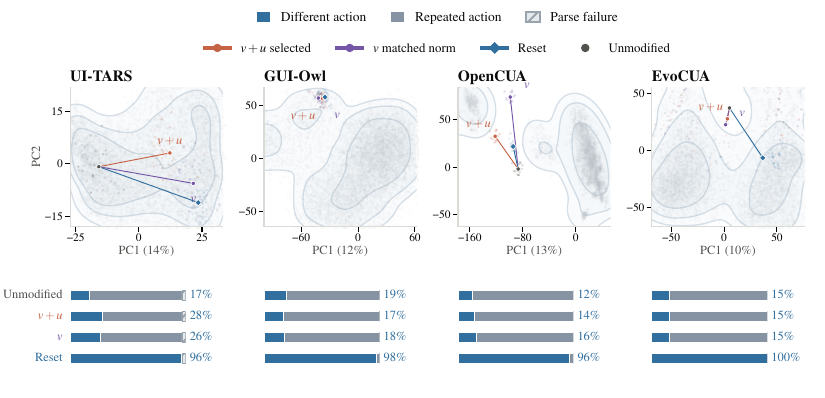}
\caption{\textbf{Activation movement does not always change the action.} Task-weighted PCA of OSWorld reference states (gray) and OpenComputer interventions (colored). Lines join the unmodified mean to composite steering, matched-norm recurrence-only steering, and reset. Bars show all proposal outcomes; percentages count proposal escapes, not executed progress.}
\label{fig:activation-movement}
\end{figure}

Context reset achieves proposal escape rates of 96\%, 98\%, 96\%, and 100\% for UI-TARS, GUI-Owl, OpenCUA, and EvoCUA. At the fixed strengths, composite steering reaches 28\%, 17\%, 14\%, and 15\%; the recurrence direction alone, scaled by $\sqrt{2}$ to match the composite per-layer norm, reaches 26\%, 18\%, 16\%, and 15\% (Figure~\ref{fig:activation-movement}). Figure~\ref{fig:activation-movement} shows that activation movement does not consistently change the proposed action.
For EvoCUA, steering leaves the replay outcome unchanged, with 84.8\% repeated proposals in the unmodified and both steered conditions.

\paragraph{Selective-reset replay.}
At the same first-trigger states, selective reset uses the history edit defined in Appendix~\ref{app:steering}, with the trajectory policy's explicit failure note and progress guard.
We regenerate two reasoning samples from the edited context and sample two actions per reasoning prefix, using the unmodified replay's seeds, generation settings, current screenshot, and action parser.
The resulting repeated-proposal rates are 70.2\%, 60.7\%, 62.5\%, and 73.9\%, respectively (Table~\ref{tab:selective-replay-intervals}).
\begin{table}[!ht]
\centering\small
\caption{Selective-reset replay outcomes (\%). Subscripts show 95\% task-bootstrap intervals. All draws enter the denominator.}
\label{tab:selective-replay-intervals}
\begin{tabular}{@{}lrrr@{}}
\toprule
Model & Repeated & Different & Parse failure\\
\midrule
UI-TARS & $70.2_{[56.5,\,82.3]}$ & $26.6_{[14.5,\,39.5]}$ & $3.2_{[0.0,\,9.7]}$\\
GUI-Owl & $60.7_{[42.9,\,77.4]}$ & $39.3_{[22.6,\,57.1]}$ & $0.0_{[0.0,\,0.0]}$\\
OpenCUA & $62.5_{[35.7,\,85.7]}$ & $37.5_{[14.3,\,64.3]}$ & $0.0_{[0.0,\,0.0]}$\\
EvoCUA & $73.9_{[58.7,\,87.0]}$ & $26.1_{[13.0,\,41.3]}$ & $0.0_{[0.0,\,0.0]}$\\
\bottomrule
\end{tabular}

\end{table}
\par\bigskip
\Needspace{6\baselineskip}
\section{Limitations}
\label{sec:limitations}

Exact recurrence certifies repeated screen--action transitions, not the absence of progress in hidden application state. The detector misses recurrences with equivalent screen content or action effects, and coordinate sensitivity differs across models (Appendix~\ref{app:tolerance}).
Failure acknowledgment concerns exposed reasoning, which differs in length across models. Human validation covers both labels in all model--benchmark combinations but uses one annotator and small stratified samples (Appendix~\ref{app:judges}).
Layer and reset-threshold selection use the reported evaluation data.
Return measurements depend on observed horizons, and activation plots project across benchmarks.
Reset's effects on actions do not establish causal mediation by the fitted directions.
History removal is not isolated from resampling and termination. Search attempts execute without rollback and can have irreversible effects.
Screen change does not establish task progress, and the task-score intervals do not establish improved completion or non-inferiority. Full-task mitigation is evaluated on OpenComputer, so transfer to other desktop or mobile environments is not established.


\begin{thebibliography}{38}
\providecommand{\natexlab}[1]{#1}
\providecommand{\url}[1]{\texttt{#1}}
\expandafter\ifx\csname urlstyle\endcsname\relax
  \providecommand{\doi}[1]{doi: #1}\else
  \providecommand{\doi}{doi: \begingroup \urlstyle{rm}\Url}\fi

\bibitem[Arditi et~al.(2024)Arditi, Obeso, Syed, Paleka, Panickssery, Gurnee,
  and Nanda]{arditi2024refusal}
Andy Arditi, Oscar Obeso, Aaquib Syed, Daniel Paleka, Nina Panickssery, Wes
  Gurnee, and Neel Nanda.
\newblock Refusal in language models is mediated by a single direction.
\newblock In \emph{Advances in Neural Information Processing Systems},
  volume~37, pp.\  136037--136083, 2024.
\newblock \doi{10.52202/079017-4322}.
\newblock URL
  \url{https://proceedings.neurips.cc/paper_files/paper/2024/hash/f545448535dfde4f9786555403ab7c49-Abstract-Conference.html}.

\bibitem[Barke et~al.(2026)Barke, Goyal, Khare, Singh, Nath, and
  Bansal]{barke2026agentrx}
Shraddha Barke, Arnav Goyal, Alind Khare, Avaljot Singh, Suman Nath, and Chetan
  Bansal.
\newblock {AgentRx}: Diagnosing {AI} agent failures from execution
  trajectories.
\newblock \emph{arXiv preprint arXiv:2602.02475}, 2026.
\newblock \doi{10.48550/arXiv.2602.02475}.
\newblock URL \url{https://arxiv.org/abs/2602.02475}.
\newblock Accepted to Findings of the Association for Computational
  Linguistics: EMNLP 2026.

\bibitem[Cemri et~al.(2025)Cemri, Pan, Yang, Agrawal, Chopra, Tiwari, Keutzer,
  Parameswaran, Klein, Ramchandran, Zaharia, Gonzalez, and
  Stoica]{cemri2025mast}
Mert Cemri, Melissa~Z. Pan, Shuyi Yang, Lakshya~A. Agrawal, Bhavya Chopra,
  Rishabh Tiwari, Kurt Keutzer, Aditya Parameswaran, Dan Klein, Kannan
  Ramchandran, Matei~A. Zaharia, Joseph~E. Gonzalez, and Ion Stoica.
\newblock Why do multi-agent {LLM} systems fail?
\newblock In \emph{Advances in Neural Information Processing Systems},
  volume~38, 2025.
\newblock \doi{10.52202/085713-4082}.
\newblock URL
  \url{https://proceedings.neurips.cc/paper_files/paper/2025/hash/b1041e52d3be19f0a9bc491657488e4a-Abstract-Datasets_and_Benchmarks_Track.html}.

\bibitem[Chen et~al.(2025)Chen, Ji, Zhong, Zhu, Li, Gan, Huang, Zou, Liu, Chen,
  Chen, and Shen]{chen2025guiShepherd}
Cong Chen, Kaixiang Ji, Hao Zhong, Muzhi Zhu, Anzhou Li, Guo Gan, Ziyuan Huang,
  Cheng Zou, Jiajia Liu, Jingdong Chen, Hao Chen, and Chunhua Shen.
\newblock {GUI-Shepherd}: Reliable process reward and verification for
  long-sequence {GUI} tasks.
\newblock \emph{arXiv preprint arXiv:2509.23738}, 2025.
\newblock \doi{10.48550/arXiv.2509.23738}.
\newblock URL \url{https://arxiv.org/abs/2509.23738}.

\bibitem[Cuadron et~al.(2025)Cuadron, Li, Ma, Wang, Wang, Zhuang, Liu,
  Schroeder, Xia, Mao, Thumiger, Desai, Stoica, Klimovic, Neubig, and
  Gonzalez]{cuadron2025dangerOverthinking}
Alejandro Cuadron, Dacheng Li, Wenjie Ma, Xingyao Wang, Yichuan Wang, Siyuan
  Zhuang, Shu Liu, Luis~Gaspar Schroeder, Tian Xia, Huanzhi Mao, Nicholas
  Thumiger, Aditya Desai, Ion Stoica, Ana Klimovic, Graham Neubig, and
  Joseph~E. Gonzalez.
\newblock The danger of overthinking: Examining the reasoning-action dilemma in
  agentic tasks.
\newblock \emph{arXiv preprint arXiv:2502.08235}, 2025.
\newblock \doi{10.48550/arXiv.2502.08235}.
\newblock URL \url{https://arxiv.org/abs/2502.08235}.

\bibitem[Elhage et~al.(2021)Elhage, Nanda, Olsson, Henighan, Joseph, Mann,
  Askell, Bai, Chen, Conerly, DasSarma, Drain, Ganguli, Hatfield-Dodds,
  Hernandez, Jones, Kernion, Lovitt, Ndousse, Amodei, Brown, Clark, Kaplan,
  McCandlish, and Olah]{elhage2021transformerCircuits}
Nelson Elhage, Neel Nanda, Catherine Olsson, Tom Henighan, Nicholas Joseph, Ben
  Mann, Amanda Askell, Yuntao Bai, Anna Chen, Tom Conerly, Nova DasSarma, Dawn
  Drain, Deep Ganguli, Zac Hatfield-Dodds, Danny Hernandez, Andy Jones, Jackson
  Kernion, Liane Lovitt, Kamal Ndousse, Dario Amodei, Tom Brown, Jack Clark,
  Jared Kaplan, Sam McCandlish, and Chris Olah.
\newblock A mathematical framework for transformer circuits.
\newblock Transformer Circuits Thread, 2021.
\newblock URL \url{https://transformer-circuits.pub/2021/framework/index.html}.
\newblock Anthropic, 22 December 2021.

\bibitem[Fawcett(2006)]{fawcett2006roc}
Tom Fawcett.
\newblock An introduction to {ROC} analysis.
\newblock \emph{Pattern Recognition Letters}, 27\penalty0 (8):\penalty0
  861--874, 2006.
\newblock \doi{10.1016/j.patrec.2005.10.010}.
\newblock URL \url{https://doi.org/10.1016/j.patrec.2005.10.010}.

\bibitem[Fu et~al.(2021)Fu, Lam, So, and Shi]{fu2021repetitionTheory}
Zihao Fu, Wai Lam, Anthony Man-Cho So, and Bei Shi.
\newblock A theoretical analysis of the repetition problem in text generation.
\newblock In \emph{Proceedings of the AAAI Conference on Artificial
  Intelligence}, volume~35, pp.\  12848--12856, 2021.
\newblock \doi{10.1609/aaai.v35i14.17520}.
\newblock URL \url{https://doi.org/10.1609/aaai.v35i14.17520}.

\bibitem[Han et~al.(2026)Han, Tu, Wang, Dai, Zhou, Lau, Cardenas, Xu, Xu,
  Xiong, Zheng, Yao, Zhou, and Xie]{han2026vlaaGui}
Qijun Han, Haoqin Tu, Zijun Wang, Haoyue Dai, Yiyang Zhou, Nancy Lau, Alvaro~A.
  Cardenas, Yuhui Xu, Ran Xu, Caiming Xiong, Zeyu Zheng, Huaxiu Yao, Yuyin
  Zhou, and Cihang Xie.
\newblock {VLAA-GUI}: Knowing when to stop, recover, and search, a modular
  framework for {GUI} automation.
\newblock \emph{arXiv preprint arXiv:2604.21375}, 2026.
\newblock \doi{10.48550/arXiv.2604.21375}.
\newblock URL \url{https://arxiv.org/abs/2604.21375}.
\newblock Version 2, 24 April 2026.

\bibitem[Holt et~al.(2024)Holt, Ruiz~Luyten, and van~der Schaar]{holt2024l2mac}
Samuel Holt, Max Ruiz~Luyten, and Mihaela van~der Schaar.
\newblock {L2MAC}: Large language model automatic computer for extensive code
  generation.
\newblock In \emph{International Conference on Learning Representations}, pp.\
  36762--36822, 2024.
\newblock URL
  \url{https://proceedings.iclr.cc/paper_files/paper/2024/hash/9e74900c3f6100c56add4bf417547848-Abstract-Conference.html}.
\newblock Appendix F: temperature annealing triggered by exact message
  recurrence.

\bibitem[Holtzman et~al.(2020)Holtzman, Buys, Du, Forbes, and
  Choi]{holtzman2020degeneration}
Ari Holtzman, Jan Buys, Li~Du, Maxwell Forbes, and Yejin Choi.
\newblock The curious case of neural text degeneration.
\newblock In \emph{International Conference on Learning Representations}, 2020.
\newblock URL \url{https://openreview.net/forum?id=rygGQyrFvH}.

\bibitem[Hu et~al.(2026)Hu, Yang, Zhou, Liang, Guo, Yin, and
  Han]{hu2026redundancyBench}
Minyang Hu, Bo~Yang, Zhinuo Zhou, Jiachen Liang, Jiahao Guo, Yiyang Yin, and
  Xiongwei Han.
\newblock Redundant or necessary? {A} benchmark for detecting redundant steps
  in agent trajectories.
\newblock \emph{arXiv preprint arXiv:2605.29893}, 2026.
\newblock \doi{10.48550/arXiv.2605.29893}.
\newblock URL \url{https://arxiv.org/abs/2605.29893}.

\bibitem[Kumar et~al.(2025)Kumar, Roh, Naseh, Karpinska, Iyyer, Houmansadr, and
  Bagdasarian]{kumar2025overthink}
Abhinav Kumar, Jaechul Roh, Ali Naseh, Marzena Karpinska, Mohit Iyyer, Amir
  Houmansadr, and Eugene Bagdasarian.
\newblock {OverThink}: Slowdown attacks on reasoning {LLM}s.
\newblock \emph{arXiv preprint arXiv:2502.02542}, 2025.
\newblock \doi{10.48550/arXiv.2502.02542}.
\newblock URL \url{https://arxiv.org/abs/2502.02542}.

\bibitem[Lee et~al.(2026)Lee, Jang, Choi, Kim, and
  Choi]{lee2026overthinkingLoops}
Yohan Lee, Jisoo Jang, Seoyeon Choi, Sangyeop Kim, and Seungtaek Choi.
\newblock Overthinking loops in agents: A structural risk via {MCP} tools.
\newblock \emph{arXiv preprint arXiv:2602.14798}, 2026.
\newblock \doi{10.48550/arXiv.2602.14798}.
\newblock URL \url{https://arxiv.org/abs/2602.14798}.

\bibitem[Madaan et~al.(2023)Madaan, Tandon, Gupta, Hallinan, Gao, Wiegreffe,
  Alon, Dziri, Prabhumoye, Yang, Gupta, Majumder, Hermann, Welleck,
  Yazdanbakhsh, and Clark]{madaan2023selfRefine}
Aman Madaan, Niket Tandon, Prakhar Gupta, Skyler Hallinan, Luyu Gao, Sarah
  Wiegreffe, Uri Alon, Nouha Dziri, Shrimai Prabhumoye, Yiming Yang, Shashank
  Gupta, Bodhisattwa~Prasad Majumder, Katherine Hermann, Sean Welleck, Amir
  Yazdanbakhsh, and Peter Clark.
\newblock {Self-Refine}: Iterative refinement with self-feedback.
\newblock In \emph{Advances in Neural Information Processing Systems},
  volume~36, pp.\  46534--46594, 2023.
\newblock \doi{10.52202/075280-2019}.
\newblock URL
  \url{https://proceedings.neurips.cc/paper_files/paper/2023/hash/91edff07232fb1b55a505a9e9f6c0ff3-Abstract-Conference.html}.

\bibitem[Marks \& Tegmark(2024)Marks and Tegmark]{marks2024geometryTruth}
Samuel Marks and Max Tegmark.
\newblock The geometry of truth: Emergent linear structure in large language
  model representations of true/false datasets.
\newblock In \emph{First Conference on Language Modeling}, 2024.
\newblock URL \url{https://arxiv.org/abs/2310.06824}.
\newblock COLM 2024; also arXiv:2310.06824v3.

\bibitem[Olsson et~al.(2022)Olsson, Elhage, Nanda, Joseph, DasSarma, Henighan,
  Mann, Askell, Bai, Chen, Conerly, Drain, Ganguli, Hatfield-Dodds, Hernandez,
  Johnston, Jones, Kernion, Lovitt, Ndousse, Amodei, Brown, Clark, Kaplan,
  McCandlish, and Olah]{olsson2022inductionHeads}
Catherine Olsson, Nelson Elhage, Neel Nanda, Nicholas Joseph, Nova DasSarma,
  Tom Henighan, Ben Mann, Amanda Askell, Yuntao Bai, Anna Chen, Tom Conerly,
  Dawn Drain, Deep Ganguli, Zac Hatfield-Dodds, Danny Hernandez, Scott
  Johnston, Andy Jones, Jackson Kernion, Liane Lovitt, Kamal Ndousse, Dario
  Amodei, Tom Brown, Jack Clark, Jared Kaplan, Sam McCandlish, and Chris Olah.
\newblock In-context learning and induction heads.
\newblock Transformer Circuits Thread, 2022.
\newblock URL \url{https://arxiv.org/abs/2209.11895}.

\bibitem[Parekh et~al.(2025)Parekh, Khayatan, Shukor, Dapogny, Newson, and
  Cord]{parekh2025learningToSteer}
Jayneel Parekh, Pegah Khayatan, Mustafa Shukor, Arnaud Dapogny, Alasdair
  Newson, and Matthieu Cord.
\newblock Learning to steer: Input-dependent steering for multimodal {LLM}s.
\newblock In \emph{Advances in Neural Information Processing Systems},
  volume~38, pp.\  159799--159834, 2025.
\newblock \doi{10.52202/085713-5341}.
\newblock URL
  \url{https://proceedings.neurips.cc/paper_files/paper/2025/hash/ea491e2d1c46686b8db5cd11154f5d2c-Abstract-Conference.html}.
\newblock NeurIPS 2025 Main Conference Track; also arXiv:2508.12815v2.

\bibitem[Qin et~al.(2025)Qin, Ye, Fang, Wang, Liang, Tian, Zhang, Li, Li,
  Huang, Zhong, Li, Yang, Miao, Lin, Liu, Jiang, Ma, Li, Xiao, Cai, Li, Zheng,
  Jin, Li, Zhou, Wang, Chen, Li, Yang, Liu, Lin, Peng, Liu, and
  Shi]{qin2025uitars}
Yujia Qin, Yining Ye, Junjie Fang, Haoming Wang, Shihao Liang, Shizuo Tian,
  Junda Zhang, Jiahao Li, Yunxin Li, Shijue Huang, Wanjun Zhong, Kuanye Li,
  Jiale Yang, Yu~Miao, Woyu Lin, Longxiang Liu, Xu~Jiang, Qianli Ma, Jingyu Li,
  Xiaojun Xiao, Kai Cai, Chuang Li, Yaowei Zheng, Chaolin Jin, Chen Li, Xiao
  Zhou, Minchao Wang, Haoli Chen, Zhaojian Li, Haihua Yang, Haifeng Liu, Feng
  Lin, Tao Peng, Xin Liu, and Guang Shi.
\newblock {UI-TARS}: Pioneering automated {GUI} interaction with native agents.
\newblock \emph{arXiv preprint arXiv:2501.12326}, 2025.
\newblock \doi{10.48550/arXiv.2501.12326}.
\newblock URL \url{https://arxiv.org/abs/2501.12326}.
\newblock UI-TARS-1.5-7B release:
  \url{https://huggingface.co/ByteDance-Seed/UI-TARS-1.5-7B}.

\bibitem[Rahn et~al.(2024)Rahn, D'Oro, and Bellemare]{rahn2024entropicSteering}
Nate Rahn, Pierluca D'Oro, and Marc~G. Bellemare.
\newblock Controlling large language model agents with entropic activation
  steering.
\newblock \emph{arXiv preprint arXiv:2406.00244}, 2024.
\newblock \doi{10.48550/arXiv.2406.00244}.
\newblock URL \url{https://arxiv.org/abs/2406.00244v2}.
\newblock Presented at the ICML 2024 Workshop on Mechanistic Interpretability;
  citing arXiv version 2, 10 October 2024.

\bibitem[Rimsky et~al.(2024)Rimsky, Gabrieli, Schulz, Tong, Hubinger, and
  Turner]{rimsky2024steering}
Nina Rimsky, Nick Gabrieli, Julian Schulz, Meg Tong, Evan Hubinger, and
  Alexander Turner.
\newblock Steering {Llama 2} via contrastive activation addition.
\newblock In \emph{Proceedings of the 62nd Annual Meeting of the Association
  for Computational Linguistics (Volume 1: Long Papers)}, pp.\  15504--15522.
  Association for Computational Linguistics, 2024.
\newblock \doi{10.18653/v1/2024.acl-long.828}.
\newblock URL \url{https://aclanthology.org/2024.acl-long.828/}.

\bibitem[Shinn et~al.(2023)Shinn, Cassano, Gopinath, Narasimhan, and
  Yao]{shinn2023reflexion}
Noah Shinn, Federico Cassano, Ashwin Gopinath, Karthik Narasimhan, and Shunyu
  Yao.
\newblock {Reflexion}: Language agents with verbal reinforcement learning.
\newblock In \emph{Advances in Neural Information Processing Systems},
  volume~36, pp.\  8634--8652, 2023.
\newblock \doi{10.52202/075280-0377}.
\newblock URL
  \url{https://proceedings.neurips.cc/paper_files/paper/2023/hash/1b44b878bb782e6954cd888628510e90-Abstract-Conference.html}.

\bibitem[Sui et~al.(2026)Sui, Chen, Li, Jiang, He, Dong, He, Gao, and
  Hooi]{sui2026tact}
Yuan Sui, Yulin Chen, Yibo Li, Xue Jiang, Yufei He, Yihong Dong, Xiaoxin He,
  Tianyu Gao, and Bryan Hooi.
\newblock {TACT}: Mitigating overthinking and overacting in coding agents via
  activation steering.
\newblock \emph{arXiv preprint arXiv:2605.05980}, 2026.
\newblock \doi{10.48550/arXiv.2605.05980}.
\newblock URL \url{https://arxiv.org/abs/2605.05980}.
\newblock Version 1, 7 May 2026; work in progress.

\bibitem[Wang et~al.(2025)Wang, Wang, Lu, Yang, Xie, Wang, Deng, Guo, Xu, Wu,
  Shen, Li, Li, Li, Chen, Zheng, Li, Lei, Cao, Fu, Shin, Shin, Hu, Wang, Chen,
  Ye, Zhang, Wang, Wang, Yang, Zhong, {Y. Charles}, Yang, and
  Yu]{wang2025opencua}
Xinyuan Wang, Bowen Wang, Dunjie Lu, Junlin Yang, Tianbao Xie, Junli Wang,
  Jiaqi Deng, Xiaole Guo, Yiheng Xu, Chen~Henry Wu, Zhennan Shen, Zhuokai Li,
  Ryan Li, Xiaochuan Li, Junda Chen, Boyuan Zheng, Peihang Li, Fangyu Lei,
  Ruisheng Cao, Yeqiao Fu, Dongchan Shin, Martin Shin, Jiarui Hu, Yuyan Wang,
  Jixuan Chen, Yuxiao Ye, Danyang Zhang, Yipu Wang, Heng Wang, Diyi Yang,
  Victor Zhong, {Y. Charles}, Zhilin Yang, and Tao Yu.
\newblock {OpenCUA}: Open foundations for computer-use agents.
\newblock In \emph{Advances in Neural Information Processing Systems},
  volume~38, pp.\  139756--139806. Curran Associates, Inc., 2025.
\newblock \doi{10.52202/085713-4669}.
\newblock URL
  \url{https://proceedings.neurips.cc/paper_files/paper/2025/hash/cc7ae529e945226b0d52ea4ac478c4f3-Abstract-Conference.html}.
\newblock OpenCUA-32B release:
  \url{https://huggingface.co/xlangai/OpenCUA-32B}.

\bibitem[Wei et~al.(2026)Wei, Ma, Zhao, Zhou, Ni, Gan, and
  Cohan]{wei2026opencomputer}
Jinbiao Wei, Qianran Ma, Yilun Zhao, Xiao Zhou, Kangqi Ni, Guo Gan, and Arman
  Cohan.
\newblock {OpenComputer}: Verifiable software worlds for computer-use agents.
\newblock \emph{arXiv preprint arXiv:2605.19769}, 2026.
\newblock \doi{10.48550/arXiv.2605.19769}.
\newblock URL \url{https://arxiv.org/abs/2605.19769}.

\bibitem[Welleck et~al.(2020)Welleck, Kulikov, Roller, Dinan, Cho, and
  Weston]{welleck2020unlikelihood}
Sean Welleck, Ilia Kulikov, Stephen Roller, Emily Dinan, Kyunghyun Cho, and
  Jason Weston.
\newblock Neural text generation with unlikelihood training.
\newblock In \emph{International Conference on Learning Representations}, 2020.
\newblock URL \url{https://openreview.net/forum?id=SJeYe0NtvH}.

\bibitem[Wollschl{\"a}ger et~al.(2025)Wollschl{\"a}ger, Elstner, Geisler,
  Cohen-Addad, G{\"u}nnemann, and Gasteiger]{wollschlager2025geometryRefusal}
Tom Wollschl{\"a}ger, Jannes Elstner, Simon Geisler, Vincent Cohen-Addad,
  Stephan G{\"u}nnemann, and Johannes Gasteiger.
\newblock The geometry of refusal in large language models: Concept cones and
  representational independence.
\newblock In \emph{Proceedings of the 42nd International Conference on Machine
  Learning}, volume 267 of \emph{Proceedings of Machine Learning Research},
  pp.\  66945--66970. PMLR, 2025.
\newblock URL \url{https://proceedings.mlr.press/v267/wollschlager25a.html}.

\bibitem[Xi et~al.(2026)Xi, Liao, Li, Zhang, Chen, Wang, Jin, Zhou, Guan, Wu,
  Ji, Gui, Zhang, and Huang]{xi2025agentprm}
Zhiheng Xi, Chenyang Liao, Guanyu Li, Zhihao Zhang, Wenxiang Chen, Binghai
  Wang, Senjie Jin, Yuhao Zhou, Jian Guan, Wei Wu, Tao Ji, Tao Gui, Qi~Zhang,
  and Xuanjing Huang.
\newblock {AgentPRM}: Process reward models for {LLM} agents via step-wise
  promise and progress.
\newblock In \emph{Proceedings of the ACM Web Conference 2026}, pp.\
  4184--4195. Association for Computing Machinery, 2026.
\newblock \doi{10.1145/3774904.3792551}.
\newblock URL \url{https://doi.org/10.1145/3774904.3792551}.

\bibitem[Xie et~al.(2024)Xie, Zhang, Chen, Li, Zhao, Cao, Hua, Cheng, Shin,
  Lei, et~al.]{xie2024osworld}
Tianbao Xie, Danyang Zhang, Jixuan Chen, Xiaochuan Li, Siheng Zhao, Ruisheng
  Cao, Toh~J Hua, Zhoujun Cheng, Dongchan Shin, Fangyu Lei, et~al.
\newblock {OSWorld}: Benchmarking multimodal agents for open-ended tasks in
  real computer environments.
\newblock \emph{Advances in Neural Information Processing Systems},
  37:\penalty0 52040--52094, 2024.
\newblock \doi{10.52202/079017-1650}.
\newblock URL
  \url{https://proceedings.neurips.cc/paper_files/paper/2024/file/5d413e48f84dc61244b6be550f1cd8f5-Paper-Datasets_and_Benchmarks_Track.pdf}.

\bibitem[Xiong et~al.(2025)Xiong, Hu, Chen, Liu, Wu, Gao, Liu, Luan, and
  Zhang]{xiong2025guiPra}
Tao Xiong, Xavier Hu, Yurun Chen, Yuhang Liu, Changqiao Wu, Pengzhi Gao, Wei
  Liu, Jian Luan, and Shengyu Zhang.
\newblock {GUI-PRA}: Process reward agent for {GUI} tasks.
\newblock \emph{arXiv preprint arXiv:2509.23263}, 2025.
\newblock \doi{10.48550/arXiv.2509.23263}.
\newblock URL \url{https://arxiv.org/abs/2509.23263}.

\bibitem[Xu et~al.(2026)Xu, Zhang, Liu, Wang, Zhu, Zhou, Hu, Gao, Cao, Wang,
  Chen, Liao, Zheng, Zeng, Xu, Bai, Lin, Zhou, and Yan]{xu2026mobileAgentV35}
Haiyang Xu, Xi~Zhang, Haowei Liu, Junyang Wang, Zhaozai Zhu, Shengjie Zhou,
  Xuhao Hu, Feiyu Gao, Junjie Cao, Zihua Wang, Zhiyuan Chen, Jitong Liao,
  Qi~Zheng, Jiahui Zeng, Ze~Xu, Shuai Bai, Junyang Lin, Jingren Zhou, and Ming
  Yan.
\newblock {Mobile-Agent-v3.5}: Multi-platform fundamental {GUI} agents.
\newblock \emph{arXiv preprint arXiv:2602.16855}, 2026.
\newblock \doi{10.48550/arXiv.2602.16855}.
\newblock URL \url{https://arxiv.org/abs/2602.16855}.
\newblock Introduces the GUI-Owl-1.5 model family.

\bibitem[Xu et~al.(2022)Xu, Liu, Yan, Cai, Li, and Li]{xu2022breakTheLoop}
Jin Xu, Xiaojiang Liu, Jianhao Yan, Deng Cai, Huayang Li, and Jian Li.
\newblock Learning to break the loop: Analyzing and mitigating repetitions for
  neural text generation.
\newblock In \emph{Advances in Neural Information Processing Systems},
  volume~35, pp.\  3082--3095, 2022.
\newblock \doi{10.52202/068431-0223}.
\newblock URL
  \url{https://proceedings.neurips.cc/paper_files/paper/2022/hash/148c0aeea1c5da82f4fa86a09d4190da-Abstract-Conference.html}.

\bibitem[Xue et~al.(2026)Xue, Peng, Huang, Guo, Han, Wang, Wang, Zhang, Yang,
  Zhao, Ding, Ma, Xie, Pei, Cai, and Qiu]{xue2026evocua}
Taofeng Xue, Chong Peng, Mianqiu Huang, Linsen Guo, Tiancheng Han, Haozhe Wang,
  Jianing Wang, Xiaocheng Zhang, Xin Yang, Dengchang Zhao, Jinrui Ding, Xiandi
  Ma, Yuchen Xie, Peng Pei, Xunliang Cai, and Xipeng Qiu.
\newblock {EvoCUA}: Evolving computer use agents via learning from scalable
  synthetic experience.
\newblock \emph{arXiv preprint arXiv:2601.15876}, 2026.
\newblock \doi{10.48550/arXiv.2601.15876}.
\newblock URL \url{https://arxiv.org/abs/2601.15876}.

\bibitem[Yang(2026)]{arora2026whyRetryingFails}
Zhanfu Yang.
\newblock Why retrying fails: Context contamination in {LLM} agent pipelines.
\newblock \emph{arXiv preprint arXiv:2605.08563}, 2026.
\newblock \doi{10.48550/arXiv.2605.08563}.
\newblock URL \url{https://arxiv.org/abs/2605.08563}.

\bibitem[Zhang et~al.(2025)Zhang, Zhang, Yang, Zhu, Zhao, Cao, Chen, and
  Yu]{zhang2025progrm}
Danyang Zhang, Situo Zhang, Ziyue Yang, Zichen Zhu, Zihan Zhao, Ruisheng Cao,
  Lu~Chen, and Kai Yu.
\newblock {ProgRM}: Build better {GUI} agents with progress rewards.
\newblock \emph{arXiv preprint arXiv:2505.18121}, 2025.
\newblock \doi{10.48550/arXiv.2505.18121}.
\newblock URL \url{https://arxiv.org/abs/2505.18121}.

\bibitem[Zhang et~al.(2026)Zhang, Xue, Wu, Chen, Liu, He, Shao, Liu, Xu, Pan,
  and Wang]{zhang2026dontActBlindly}
Yuzhe Zhang, Xianwei Xue, Xingyong Wu, Mengke Chen, Chen Liu, Xinran He, Run
  Shao, Feiran Liu, Huanmin Xu, Qiutong Pan, and Haiwei Wang.
\newblock Don't act blindly: Robust {GUI} automation via action-effect
  verification and self-correction.
\newblock In Maria Liakata, Viviane~P. Moreira, Jiajun Zhang, and David Jurgens
  (eds.), \emph{Proceedings of the 64th Annual Meeting of the Association for
  Computational Linguistics (Volume 1: Long Papers)}, pp.\  28924--28941, San
  Diego, California, United States, July 2026. Association for Computational
  Linguistics.
\newblock ISBN 979-8-89176-390-6.
\newblock \doi{10.18653/v1/2026.acl-long.1335}.
\newblock URL \url{https://aclanthology.org/2026.acl-long.1335/}.
\newblock System name VeriGUI; also arXiv:2604.05477v1.

\bibitem[Zheng et~al.(2023)Zheng, Chiang, Sheng, Zhuang, Wu, Zhuang, Lin, Li,
  Li, Xing, Zhang, Gonzalez, and Stoica]{zheng2023llmJudge}
Lianmin Zheng, Wei-Lin Chiang, Ying Sheng, Siyuan Zhuang, Zhanghao Wu, Yonghao
  Zhuang, Zi~Lin, Zhuohan Li, Dacheng Li, Eric~P. Xing, Hao Zhang, Joseph~E.
  Gonzalez, and Ion Stoica.
\newblock Judging {LLM}-as-a-judge with {MT-Bench} and {Chatbot Arena}.
\newblock In \emph{Advances in Neural Information Processing Systems},
  volume~36, pp.\  46595--46623, 2023.
\newblock \doi{10.52202/075280-2020}.
\newblock URL
  \url{https://papers.neurips.cc/paper_files/paper/2023/hash/91f18a1287b398d378ef22505bf41832-Abstract-Datasets_and_Benchmarks.html}.

\bibitem[Zhou et~al.(2026)Zhou, Qu, Wu, Kim, Prakash, Rus, Low, and
  Liang]{zhou2026mem1}
Zijian Zhou, Ao~Qu, Zhaoxuan Wu, Sunghwan Kim, Alok Prakash, Daniela Rus, Bryan
  Kian~Hsiang Low, and Paul~Pu Liang.
\newblock {MEM1}: Learning to synergize memory and reasoning for efficient
  long-horizon agents.
\newblock In \emph{International Conference on Learning Representations}, 2026.
\newblock URL
  \url{https://proceedings.iclr.cc/paper_files/paper/2026/hash/5fc8b3bdfbb9167b5144df5d3fae4616-Abstract-Conference.html}.

\end{thebibliography}
\end{document}